\documentclass[%
 aip,
 amsmath,amssymb,
 reprint,%
]{revtex4-1}

\usepackage{graphicx}
\usepackage{dcolumn}
\usepackage{bm}

\usepackage[utf8]{inputenc}
\usepackage[T1]{fontenc}
\usepackage{mathptmx}
\usepackage{etoolbox}
\usepackage{float}
\usepackage{mathtools}
\usepackage{nicefrac}
\usepackage[mathcal]{euscript}

\makeatletter
\def\@email#1#2{%
 \endgroup
 \patchcmd{\titleblock@produce}
  {\frontmatter@RRAPformat}
  {\frontmatter@RRAPformat{\produce@RRAP{*#1\href{mailto:#2}{#2}}}\frontmatter@RRAPformat}
  {}{}
}%
\makeatother

\DeclareMathOperator*{\argmax}{arg\,max}
\DeclareMathOperator*{\sgn}{sgn}

\begin{document}

\preprint{AIP/123-QED}

\title[Tailored minimal reservoir computing]{Tailored minimal reservoir computing: on the bidirectional connection between nonlinearities in the model and in data}
\author{Davide Prosperino}
 \affiliation{Ludwig-Maximilians-Universität München, Faculty of Physics, Geschwister-Scholl-Platz 1, 80539 Munich, Germany}
\author{Haochun Ma}
 \affiliation{Ludwig-Maximilians-Universität München, Faculty of Physics, Geschwister-Scholl-Platz 1, 80539 Munich, Germany}
\author{Christoph Räth}
 \affiliation{Deutsches Zentrum für Luft- und Raumfahrt (DLR), Institute for Frontier Materials on Earth and in Space, Linder Höhe, 51147 Cologne, Germany}
 \email{christoph.raeth@dlr.de}

\date{11 August 2025}

\begin{abstract}
We study how the degree of nonlinearity in the input data affects the optimal design of reservoir computers, focusing on how closely the model’s nonlinearity should align with that of the data.
By reducing minimal RCs to a single tunable nonlinearity parameter, we explore how the predictive performance varies with the degree of nonlinearity in the model.
To provide controlled testbeds, we generalize to the fractional Halvorsen system, a novel chaotic system with fractional exponents.
Our experiments reveal that the prediction performance is maximized when the model’s nonlinearity matches the nonlinearity present in the data.
In cases where multiple nonlinearities are present in the data, we find that the correlation dimension of the predicted signal is reconstructed correctly when the smallest nonlinearity is matched.
We use this observation to propose a method for estimating the minimal nonlinearity in unknown time series, by sweeping the model exponent and identifying the transition to a successful reconstruction.
Applying this method to both synthetic and real-world datasets, including financial time series, we demonstrate its practical viability.
Additionally, we briefly study the SINDy framework as a complementary approach for identifying nonlinearities in data.
Finally, we transfer these insights to classical RCs, by augmenting traditional architectures with fractional, generalized reservoir states.
This yields performance gains, particularly in resource-constrained scenarios, such as physical reservoirs, where increasing reservoir size is impractical or economically unviable.
Our work provides a principled route toward tailoring RCs to the intrinsic complexity of the systems they aim to model.
\end{abstract}

\maketitle

\begin{quotation}
Reservoir computing is a powerful tool for modeling nonlinear systems, but its design often relies on heuristics.
Here, we show that predictive accuracy improves when the model’s nonlinearity is matched to that of the data.
Using a deterministic, minimal reservoir framework and a novel chaotic system with tunable fractional exponents, we isolate this relationship and demonstrate that the smallest nonlinearity in the data plays a key role.
This insight enables a method to estimate nonlinearity from time series alone, which we validate on synthetic and real-world data, including financial markets.
We also show how augmenting classical reservoirs with tailored nonlinearities improves performance, which is especially useful in hardware-limited settings.
\end{quotation}

\section{\label{sec:introduction}Introduction}
Reservoir computing\cite{Jaeger2001,Maass2002,Jaeger2004} is a machine learning framework for modeling and predicting nonlinear dynamical systems, built on the idea of using a fixed recurrent dynamical system---the \textit{reservoir}---and linearly combining its dynamics to create predictions.
The work by \textcite{LukoseviciusJaeger2009} offers a great introduction to the theory of traditional reservoir computers (RCs).

Despite its practical success in synthetic systems\cite{Pathak2017, Lu2018} and real-world systems\cite{Shahi2021, Brucke2024, Herteux2024, Li2024, Mijalkov2025}, classical reservoir computers remain somewhat heuristic:
the reservoir's weights are initialized randomly, and while empirical studies on the reservoir structure and weights have been performed\cite{Carroll2019, Haluszczynski2019}, the optimal design of the reservoir is not well understood analytically.
This randomness and complexity hinder a principled understanding of \textit{why} RCs work so well, since we need to account not only for the choice of parameters, but also for the actual realization of the random numbers used in the process.

The topic of randomness in traditional RC has been addressed in two ways:
In so-called `next generation reservoir computing' (NGRC)\cite{Gauthier2021} the reservoir is replaced by linear and nonlinear combinations of the input variables and their time lags without any weights.
While this approach performs very well, also with limited training data\cite{Barbosa22}, the typical character of a reservoir as being a dynamical substrate with (fading) memory, which responds to some input data, is lost in this RC setup.
In minimal reservoir computing (minRC)\cite{Ma2023MinimalRC}, however, the reservoir still exists and the reservoir states are iteratively fed through a reservoir creating a dynamical system. 
The key difference between the two is that NGRC eliminates the reservoir entirely and replaces it with handcrafted lagged features, whereas minimal RC retains a true, though simplified, dynamical reservoir structure. As a result, minimal RC preserves the core concept of memory-driven dynamics, while NGRC trades this for a purely feature-based representation.

The reservoir in minimal RC exists, but it is simplified by replacing the large random network with a structured block-diagonal matrix splitting the reservoir into multiple smaller sub-reservoirs, each working on a single feature.
All random elements in the input layer and in the reservoir are removed, enabling a systematic analysis of RC architectures.
Furthermore, the nonlinear activation at each reservoir node can be replaced by shifting the nonlinearity to the output layer: The readout operates on generalized reservoir states that include powers of the reservoir’s linear state evolution.
This deterministic setup retains RC’s computational efficiency but yields a more interpretable, tractable model.

Building on this minimal RC framework, this work focuses on a fundamental question:
How nonlinear should the reservoir states be in order to adequately model given nonlinear input data?
In classical or minimal RC the approach is to introduce nonlinear reservoir features to help capture nonlinear structures in the input.
However, it remains unclear what degree of nonlinearity is truly needed in the reservoir states to represent the nonlinear dynamics of the data.
Intuitively, if the input data’s dynamics are only mildly nonlinear, an overly strong nonlinearity in the reservoir states might be unnecessary (or even detrimental), whereas if the data’s generative process is highly nonlinear, linear or weakly nonlinear reservoir states will be insufficient to capture its behavior.
We aim to formalize this intuition and determine how to tailor the model’s nonlinearity to the complexity of the input.

In this study, we introduce a tailored minimal RC approach to systematically investigate the matching of data and reservoir states nonlinearities.
In Sec. \ref{sec:reservoir-computing} we reduce the minimal RC model to its essence by using a single tunable nonlinearity parameter in the generalized reservoir states, and we examine the impact of this nonlinearity on prediction performance for datasets of varying complexity.

All traditional chaotic systems use integer exponents as nonlinearities with the Thomas system's\cite{Thomas1999} sine function being a notable exception.
Here in Sec. \ref{sec:data} we introduce a fractional Halvorsen system as a novel data generator, generalizing the classical Halvorsen attractor to fractional exponents in the nonlinear terms.
This allows us to produce time series with a controllable degree of nonlinearity, providing an ideal testbed for our study.

We use this data to perform extensive experiments in Sec. \ref{sec:mininal-required-nonlinearity} measuring the prediction performance and studying the relationship between nonlinearity in the data and nonlinearity in the model.

Applying our findings in reverse, we find in Sec. \ref{sec:applications} that we can use this framework to determine the smallest nonlinearity present in the data by measuring the prediction performance over different nonlinearities in the model and noting when the prediction error minimizes.
There we compare our findings against using SINDy.
Lastly, in that section we also transfer our insights from minimal RCs to improve the performance of classical RCs by introducing fractional, generalized reservoir states.

\section{\label{sec:data}Data}
Originally introduced as a model for atmospheric convection, the Lorenz system\cite{Lorenz1963} has become the benchmark system in the study of chaotic systems and has been extensively used in research on reservoir computers\cite{Pathak2017, Lu2018}.
However, its nonlinearities consist of mixed variables, which make it hard to control the exponent, and thus the nonlinearity.
For this reason, we introduce the Halvorsen system\cite{Sprott2010Halvorsen} in our study, after performing initial studies on the Lorenz system.
We introduce a modified version of the Halvorsen system, in which we can control the nonlinearity in the data more precisely.

For all our integrations of the trajectories, we use the explicit Runge--Kutta method of order 5(4)\cite{RungeKutta45} utilizing a step size of $\Delta t=0.01$ unless stated otherwise.
Our initial condition consists of a uniformly distributed random value between $-20$ and $20$ for the Lorenz system, and due to stability reasons we use the point $\begin{pmatrix}0.1&0&0\end{pmatrix}^\textrm{T}$ as the initial condition for all Halvorsen realizations.
In each case we discard the first $10^4$ steps as transient behavior.

For all calculations involving reservoir computers, we use the SCAN software package\cite{SCAN}, and for all calculations involving SINDy we use the PySINDy package\cite{pysindy1,pysindy2}.

\subsection{Lorenz system}
The Lorenz system is a set of coupled nonlinear differential equations given by\cite{Lorenz1963}
\begin{subequations}
\label{equ:lorenz-equation}
\begin{align}
\dot{x}_1&=-\sigma\,x_1 + \sigma\,x_2\\
\dot{x}_2&=\rho\,x_1 - x_2 - x_1\,x_3\\
\dot{x}_3&=-\beta\,x_3 + x_1\,x_2\;\;\;,
\end{align}
\end{subequations}
where we use the standard parametrization exhibiting chaotic behavior of $\sigma=10$, $\rho=28$, and $\beta=\nicefrac{8}{3}$.

While previous studies have explored variations in the Lorenz system’s nonlinear terms\cite{Ma2022Causality}, controlling the overall degree of nonlinearity across coordinates remains challenging due to the nonlinearity consisting of combinations of two variables.

\subsection{Fractional Halvorsen system\label{sec:data-halvorsen}}
The Halvorsen system, in contrast to the Lorenz system, is a chaotic system which has its nonlinearities in a single variable in each dimension.
Originally, the nonlinearity consists of quadratic terms.
However, in this work we want to introduce fractional exponents in each dimension.
This allows for a control of the nonlinearity of the system by modifying the exponent in each equation, which leads to the introduction of the modified, fractional Halvorsen system given by
\begin{subequations}
\label{equ:halvorsen-equation}
\begin{align}
\dot{x}_1&=-a\,x_1 - 4\,x_2 - 4\,x_3 - x_2^{\xi_1}\\
\dot{x}_2&=-a\,x_2 - 4\,x_3 - 4\,x_1 - x_3^{\xi_2}\\
\dot{x}_3&=-a\,x_3 - 4\,x_1 - 4\,x_2 - x_1^{\xi_3}\;\;\;.
\end{align}
\end{subequations}
The canonical choice displaying chaotic behavior is $a=1.3$ and $\xi_i=2$.
However, for this study we want to explore different values of $\xi_i$.
Here, we do not want to limit ourselves to integer exponents, but introduce the study of fractional exponents in this context.

For fractional exponents of the form $\xi_i = \nicefrac{n_i}{d}$ we need to rewrite Eqs. \ref{equ:halvorsen-equation} slightly and use the definition of $x^{\tfrac{n_i}{d}} = \sqrt[\leftroot{1}\uproot{1}d]{x^{\,n_i}}$ to substitute for $x^{\xi_i}$.
In order to prevent complex-valued trajectories, we limit ourselves to even choices for $n_i$.

During our studies we discovered that a denominator of $d=50$ provides a good trade-off between the accessible granularity and computational stability.

\begin{figure*}
\centering
\includegraphics[width=\textwidth]{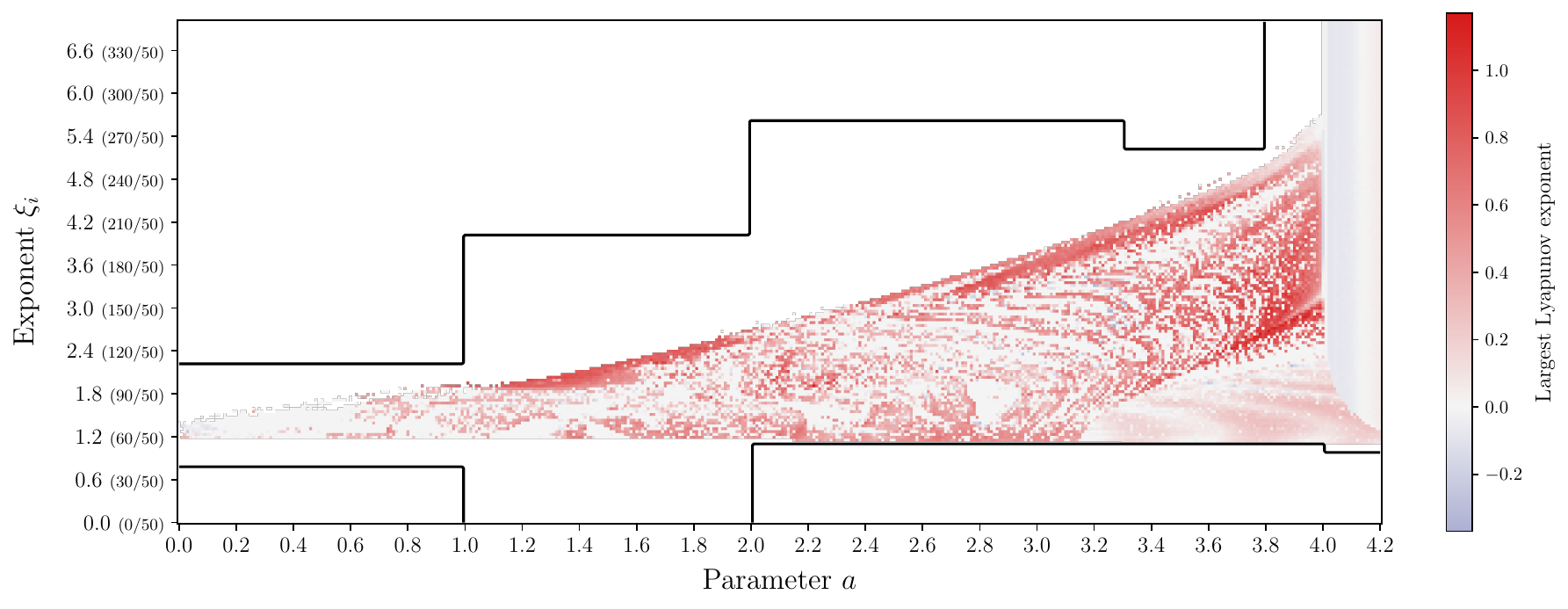}
\caption{\label{fig:lyapunov-rational-halvorsen}The results of our grid search for the calculation of the largest Lyapunov exponent for different parameters of $a$ and $\xi_i$ are shown.
For this plot all exponents of the fractional Halvorsen system are the same and fixed at $\xi_i$.
White color inside the black boundary indicates a diverging trajectory for that parameter combination, while a white color outside the boundary indicates that the parameter combination has not been explored.
For each parameter combination we simulate $50\,000$ steps and discard the first $10\,000$ steps as transient behavior.
In total, we performed $44\,625$ experiments.}
\end{figure*}

In Fig. \ref{fig:lyapunov-rational-halvorsen} we performed a grid search over the two free parameters of the fractional Halvorsen system, in order to find the region of interest with chaotic behavior.
Here, we fix all exponents to be the same with $\xi_i = \xi_1 = \xi_2 = \xi_3$.
We found that for a parameter of $a = 3.98$ we observe chaotic behavior over a range of exponents $\xi_i$.

\subsection{\label{sec:thomas-system}Thomas system}
The Thomas system\cite{Thomas1999} is a chaotic system, which has its nonlinearity not in exponentiation, but instead in a sine function.
It is defined by
\begin{subequations}
\begin{align}
    \dot{x}_1 &= -b\,x_1+\sin{x_2}\\
    \dot{x}_2 &= -b\,x_2+\sin{x_3}\\
    \dot{x}_3 &= -b\,x_3+\sin{x_1}\;\;\;,
\end{align}
\end{subequations}
where we use a parameter of $b=0.21$.
The sine function can be defined by Taylor expansion as a sum of polynomials, where the first nonlinear term appears as third order.

\subsection{Surrogate systems}
Surrogating the data destroys the nonlinear properties of a time series while keeping the linear ones unaffected\cite{Theiler1992}.
In Sec. \ref{sec:finding-smallest-nonlinearity} we aim to identify the smallest nonlinearity in a given time series.
To ensure that the observed effects indeed arise from nonlinearity in the data, we generate Fourier transform (FT) surrogates.
If the measure from the original time series differs significantly from this linearized background, it strongly indicates that the observed effect stems from the system’s nonlinearity.
In this study we restrict ourselves to the use of FT surrogates, since there is evidence that only this class of surrogates reliably destroys all nonlinearities in the data\cite{Raeth2012, Laut2016, Schreiber2018}.

We create the surrogate time series by first performing a Fourier transformation $\mathcal{F}$ on the original time series $\underline{x}$, separating the data into amplitudes $A_k$ and phases $\phi_k$.
The linear properties are now stored in the amplitudes and the nonlinear ones in the phases.
By replacing the original phases $\phi_k$ with uniformly distributed numbers between $[0,\;2\,\pi]$, $\phi_k^{\textrm{rand}}$, we destroy the nonlinear features of the original time series.
The surrogate time series $\underline{x}_{\textrm{s}}$ is then given by the inverse Fourier transformation $\mathcal{F}^{-1}$ of the original amplitudes $A_k$ with the randomized phases $\phi_k^{\textrm{rand}}$, sketched by
\begin{equation}
     \underline{x}_{\textrm{s}} = \mathcal{F}^{-1}{\left[A_k\,\exp{\left( i\,\phi_k^{\textrm{rand}} \right)}\right]}\;\;\;.
\end{equation}

In order to get a robust estimate of the surrogate measures, we create multiple realizations of the surrogate time series and calculate the measure of interest across all of them and report the average with a standard deviation indicating the spread of the values.

\section{\label{sec:reservoir-computing}Reservoir computing}
A reservoir computer (RC) is a specialized form of recurrent neural network in which the recurrent connections remain fixed after initialization, rather than being adapted during training.
The input signal is mapped into a random, high-dimensional state space, causing the randomly defined reservoir to synchronize with the input’s dynamics.
The resulting reservoir states now reflect the time evolution of the input in the high-dimensional space.
They are then combined through a linear readout layer to produce predictions in the measured space.
Such architectures have been shown to excel at forecasting chaotic systems.
For a thorough overview of classical reservoir computing techniques we refer to \textcite{LukoseviciusJaeger2009}.

\subsection{Reservoir computers}
In a classical reservoir computer, the input time series $\underline{x}$ is mapped \textit{randomly} into a high-dimensional space of dimensionality $d$ through the input matrix $\mathbf{W}_\textrm{in}$.
Once the data is embedded, the reservoir governs the internal dynamics of the reservoir computer.
The reservoir states are then linearly combined to form the prediction.

The reservoir is a \textit{randomly} connected graph of $d$ nodes represented by its adjacency matrix $\mathbf{A}$, which describes the connection of the nodes with each other.
Different topologies for the reservoir $\mathbf{A}$ have been studied, showing that, generally, random networks or small-world networks work better than scale-free networks\cite{Haluszczynski2019}.
The reservoir is scaled to a target spectral radius of $\rho^*$ to regulate the reservoir's dynamical stability and ensure that it does not diverge.
At each time $t$, we represent the reservoir by the state vector $\underline{r}{(t)}$, whose components reflect the activity of each node.
Its evolution is given by
\begin{equation}
    \underline{r}{(t+1)} = f{\left(\mathbf{A}\,\underline{r}(t) + \mathbf{W}_\textrm{in}\,\underline{x}(t)\right)}\;\;\;,
\end{equation}
and it is usually initialized with the zero vector $\underline{r}(0) = \underline{0}$.
$f$ refers to a nonlinear function and the hyperbolic tangent is the usual choice.

After a synchronization phase, sometimes referred to as the warm-up phase, the reservoir state $\underline{r}$ represents the dynamics of the input data $\underline{x}$ in the high-dimensional space.
The reservoir states can then be linearly combined by an output matrix $\mathbf{W}_\textrm{out}$ to reproduce the time series in the original space.

For finding the output matrix $\mathbf{W}_\textrm{out}$, the reservoir states $\underline{r}(t+1)$ with their corresponding output $\underline{x}(t+1)$ are recorded during the training process, collecting a total number of $l$ training steps.
We store the reservoir states and their corresponding outputs in the matrices $\mathbf{R}$ and $\mathbf{X}$ respectively, and perform a ridge regression\cite{RidgeRegression} to solve the equation $\mathbf{W}_\textrm{out}\,\mathbf{R} = \mathbf{X}$.
The solution for the output matrix is given by
\begin{equation}
    \mathbf{W}_\textrm{out} = \mathbf{X}\,\mathbf{R}^\textrm{T}\,\left( \mathbf{R}\,\mathbf{R}^\textrm{T} + \beta\,\mathbf{1}\right)^{-1}\;\;\;,
\end{equation}
where we have applied the mathematical trick described by \textcite{LukoseviciusJaeger2009}, consisting of multiplying $\mathbf{R}^{\textrm{T}}$ to the right of the problem to solve, in order to make the optimization independent of the training length.
Here, $\mathbf{1}$ describes the identity matrix, and $\beta$ the regularization parameter of the ridge regression.

The reservoir states can also be generalized before the optimization as already studied in different works\cite{Herteux2020, Ohkubo2024}.
In this work we briefly study the effects of generalizing the reservoir states to fractional powers using $\underline{\tilde{r}} = \begin{pmatrix} \underline{r} & \underline{r}^\eta \end{pmatrix}^{\textrm{T}}$.

For creating predictions after training, the reservoir computer needs to be synchronized to the immediate history of the starting point of the prediction to ensure that the reservoir state represents the current dynamics.
After that, the predictions can be fed successively into the reservoir computer to reproduce the learned dynamics.

\subsection{Minimal reservoir computers}
Classical reservoir computers utilize random initializations, rendering their study challenging since we must account for both the chosen setup and the particular realization of the random numbers.
To eliminate the element of randomness in reservoir computing, we introduced minimal reservoir computers\cite{Ma2023MinimalRC} as architectures defined entirely without random components.
This simplifies analyzing their inner workings, as the absence of random initializations and network configurations allows for a more direct examination of the reservoir’s dynamics.

Minimal reservoir computers can be seen as deterministic subsets of classical reservoir computing approaches.
In the following, we outline their definition, but want to refer to Ref. \onlinecite{Ma2023MinimalRC} for a detailed discussion.

The input data are not embedded randomly in a high-dimensional space.
Instead of creating random features from the data, as done in classical RC, for minimal RC we construct the features from the set of all subset sums of the coordinates.
We use all partial sums which are creatable by the coordinates and we feed multiple copies of each feature into the reservoir.
The number of copies fed into the reservoir is defined by the block size $b$ and each copy is assigned a weight in $[0,\;1]$ according to the weight vector $\underline{w}$ given by
\begin{equation}
    \underline{w} = \begin{pmatrix}
        1 & \sqrt{\dfrac{b-2}{b-1}} & \cdots & \sqrt{\dfrac{1}{b-1}} & 0
    \end{pmatrix}^{\textrm{T}}\;\;\;.
\end{equation}
For a three-dimensional system the input matrix $\mathbf{W}_\textrm{in}$ is constructed by
\begin{equation}
    \mathbf{W}_\textrm{in} =
    \begin{pmatrix}
        \underline{w} & \underline{0} & \underline{0}\\
        \underline{0} & \underline{w} & \underline{0}\\
        \underline{0} & \underline{0} & \underline{w}\\
        \underline{w} & \underline{w} & \underline{0}\\
        \underline{w} & \underline{0} & \underline{w}\\
        \underline{0} & \underline{w} & \underline{w}\\
        \underline{w} & \underline{w} & \underline{w}\\
    \end{pmatrix}\;\;\;,
\end{equation}
resulting in the following feature vector being fed into the reservoir:
\begin{equation}
\begin{aligned}
\mathbf{W}_\textrm{in}\,\underline{x} =
\begin{pmatrix}
        \underline{w} \odot \underline{x_1} & \underline{w} \odot \underline{x_2}
         & \cdots & \underline{w} \odot \underline{x_{1+2}}& \cdots & \underline{w} \odot \underline{x_{1+2+3}}\end{pmatrix}^{\textrm{T}}\;\;\;.
\end{aligned}
\end{equation}
Here, $\odot$ describes the element-wise multiplication between the two vectors, and for each feature $f$, the vector $\underline{x_f}$ is defined as $\underline{x_f} = \begin{pmatrix} x_f & \cdots & x_f \end{pmatrix}^\textrm{T}$ to match the dimensionality of $\underline{w}$.
The subscript indicates the coordinates out of which the feature is constructed by summing over them.

Instead of using a single, big reservoir, the reservoir is constructed as several, disconnected, smaller reservoirs, which leads to an adjacency matrix in block-diagonal form.
For each feature $f$ we use a small reservoir $\mathbf{J}_f$ consisting of a matrix of ones, meaning that each node is connected to every other node.
The final reservoir is then constructed by
\begin{equation}
    \mathbf{A} = \dfrac{\rho^*}{b}\,
    \begin{pmatrix}
        \mathbf{J}_{x_1}&\mathbf{0}&\cdots&\mathbf{0}\\
        \mathbf{0}&\mathbf{J}_{x_2}&\cdots&\mathbf{0}\\
        \vdots&\vdots&\ddots&\vdots\\
        \mathbf{0}&\mathbf{0}&\cdots&\mathbf{J}_{x_{1+2+3}}\\
    \end{pmatrix}\;\;\;,
\end{equation}
where the scaling factor of $\nicefrac{\rho^*}{b}$ ensures that the reservoir $\mathbf{A}$ has the spectral radius of $\rho^*$.
The idea of block-diagonal reservoirs has also been successfully applied to classical RC architectures\cite{Ma2023Blockdiagnoal, Li2024}.

Unlike in classical RC architectures, in minimal RC the reservoir states are evolved purely linearly by
\begin{equation}
    \underline{r}{(t+1)} = \mathbf{A}\,\underline{r}{(t)} + \mathbf{W}_\textrm{in}\,\underline{x}{(t)}\;\;\;.
\end{equation}
In the original definition, the nonlinearity is added \textit{after} the evolution by extending the reservoir state to a generalized reservoir state $\underline{\tilde{r}}$ containing copies of itself raised up to a maximal power of $\eta_\textrm{max}$ given by
\begin{equation}
    \underline{\tilde{r}} = \begin{pmatrix}
        \underline{r} & \underline{r}^2 & \cdots & \underline{r}^{\eta_\textrm{max}-1} & \underline{r}^{\eta_\textrm{max}}
    \end{pmatrix}^{\textrm{T}}\;\;\;.
\end{equation}
The exponentiation is understood to be applied element-wise.

However, in this article we want to introduce a slightly modified setup of the generalized states containing only the linear reservoir state and a single nonlinearity $\eta$.
The generalized reservoir state $\underline{\tilde{r}}$ reduces to
\begin{equation}
    \underline{\tilde{r}} = \begin{pmatrix}
        \underline{r} & \underline{r}^\eta
    \end{pmatrix}^{\textrm{T}}\;\;\;.
\end{equation}

Additionally, we also want to allow for fractional nonlinearities of the form $\eta = \nicefrac{n}{d}$, where we apply the same substitution as in Sec. \ref{sec:data-halvorsen} of $\underline{r}^{\frac{n}{d}} = \sqrt[\leftroot{1}\uproot{1}d]{\underline{r}^{\,n}}$.
Each operation is understood to be performed element-wise and again we only allow even numerators $n$ in order to prevent complex-valued reservoir states.

Utilizing this reduced definition of minimal RCs, we can study the dependence between the nonlinearity present in data and the smallest nonlinearity required to successfully predict those systems.

The training and prediction routines are identical to the classical RC's case.
We train each minimal RC by performing a ridge regression of the generalized reservoir states at each time point against the corresponding output, and create predictions by iteratively inputting the previous prediction.

\section{\label{sec:sindy}SINDy}
Sparse identification of nonlinear dynamics (SINDy)\cite{SINDy} is a method for identifying the underlying governing equations of a dynamical system using a user-defined library.
The measured trajectory of a system is collected over a number of training steps in a matrix $\mathbf{X}$.
Then, based on the library $\boldsymbol{\Theta}$ a collection of candidate terms $\boldsymbol{\Theta}{\left(\mathbf{X}\right)}$ is built.
Iterative sparse regression is then applied on this collection to identify the fewest terms that accurately represent the system’s time derivative.

The library $\boldsymbol{\Theta}$ is of key importance in the SINDy framework as it describes all possible interactions and terms.
Usually, it is built to be as large as reasonable for allowing the optimizer to find all appropriate nonlinearities to accurately describe the system at hand.
Typical choices of the library include polynomial terms up to a certain degree, trigonometric functions, and exponential functions\cite{Bhadriraju2020, Paparazzo2025}.

However, in this article we want to use SINDy differently.
Instead of providing a large selection of nonlinearities, we only provide a single nonlinearity and study the performance on the reconstruction.
Additionally, we exclusively use fractional nonlinearities $\eta$ as defined for minimal RC.

In the notation of \textcite{SINDy}, we define our library for a single nonlinearity $\eta$ as
\begin{equation}
\boldsymbol{\Theta}{\left(\mathbf{X}\right)} =
\begin{pmatrix}
    \vline & \vline \\
    \mathbf{X} & \mathbf{X}^\eta \\
    \vline & \vline
\end{pmatrix}\;\;\;.
\end{equation}

The terms for the fractional exponents $\mathbf{X}^\eta$ are constructed as follows.
First, we define a global list of all possible nonlinearities $\mathbf{H} \subset \tfrac{1}{d}\mathbb{Z}^+$.
For a single nonlinearity $\eta \in \mathbf{H}$ we then use all possible terms for $\mathbf{X}^\eta$, such that all possible permutations of coordinates, which result in the target nonlinearity $\eta$, are included.
We sketch this by
\begin{align}
\mathbf{X}^{\eta} = \left\{ \prod_{c_i \in C} c_i^{\alpha_i}\,\sgn{c_i} \;\middle|\; \right.  &C \in \mathcal{P}{\left(\begin{Bmatrix}x_1&\dots&x_n\end{Bmatrix}\right)}\setminus \{\emptyset\}, \notag \\ &\left. \;\sum_{i} \alpha_i = \eta,\; \alpha_i \in \mathbf{H}\right\}\;\;\;,
\end{align}
where $\mathcal{P}$ describes the power set and $\sgn$ the sign function.
We provide an illustration of this definition in Eq. \ref{equ:sindyexample} in the appendix.

We then solve the equation $\dot{\mathbf{X}} = \boldsymbol{\Theta}{\left(\mathbf{X}\right)}\,\boldsymbol{\Xi}$ for $\boldsymbol{\Xi}$ to obtain the weights for each term in the library.
We optimize using the sequentially thresholded least squares (STLSQ) algorithm\cite{SINDy} and the parameter $\alpha$ describes the regularization strength of the optimization in each iteration.

\section{\label{sec:mininal-required-nonlinearity}Minimal required nonlinearity}
In this section we want to present our results and analyze the connection between nonlinearities expressed in the data and the required nonlinearities for reproducing those.

\subsection{Lorenz system}
\begin{figure*}
\centering
\includegraphics[width=\textwidth]{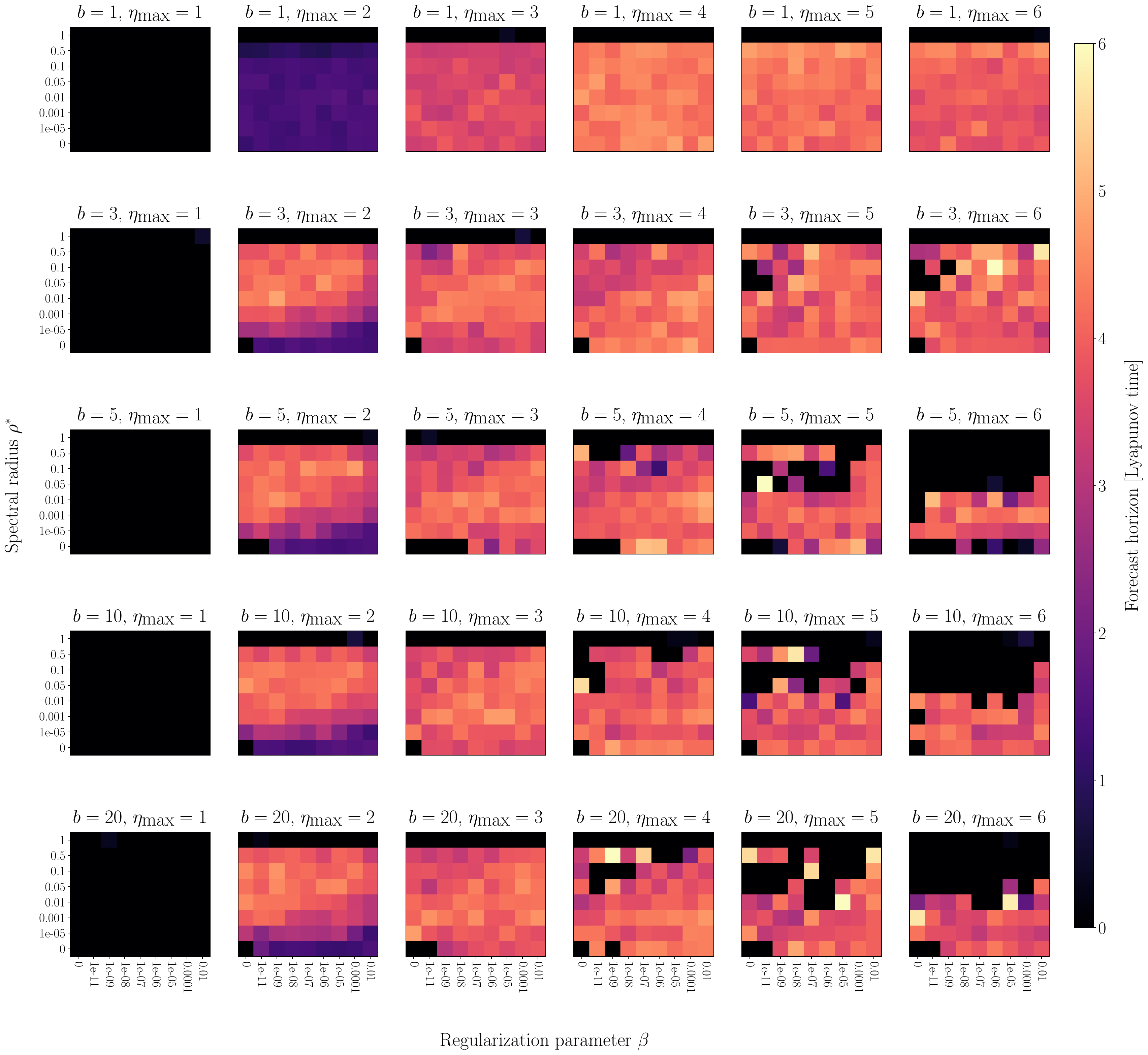}
\caption{\label{fig:minimal-rc-lorenz-big-grid}The performance for different hyperparameters of minimal RCs in the classical setup, containing all nonlinearities in the generalized states up to $\eta_\textrm{max}$ predicting the Lorenz system is shown.
The performance of successful runs using the forecast horizon is measured in multiples of Lyapunov times.
For each realization we use $1\,000$ data points for training, out of which 10 are used for synchronization, and a step size of $\Delta t=0.025$.
Each tile shows the average performance of at least 35 realizations and in total we performed $98\,297$ experiments.}
\end{figure*}

We begin our analysis with a wide grid search over \textit{all} hyperparameters of minimal RCs in Fig. \ref{fig:minimal-rc-lorenz-big-grid} for the Lorenz system.
We want to emphasize at this point that each tile in Fig. \ref{fig:minimal-rc-lorenz-big-grid} completely and uniquely describes a minimal reservoir computer instance.
For minimal RCs, the repeated experiments for a certain setup solely average out the effect of different training data or put differently, being on different parts of the attractor.
In contrast to classical RCs, where repeated experiments are required to control for the randomness in their construction in addition to different parts of the attractor being used for training.

For the results in Fig. \ref{fig:minimal-rc-lorenz-big-grid} we utilize the traditional setup of minimal RC, where all integer exponents up to a maximal exponent $\eta_\textrm{max}$ are used in the generalized reservoirs state $\underline{\tilde{r}}$.
We observe that the prediction is successful for a wide range of hyperparameters using a minimal data setup of only $1\,000$ training steps.
Additionally, we can report that the performance seems to increase when including higher order nonlinearities.
So, including higher nonlinearities, which are not present in the data, seems to give the minimal RC more flexibility and allow a more precise approximation.
Increasing the block size $b$ leads to instabilities for higher order terms, which is why we recommend using a relatively small block size not exceeding $b = 5$.

In our findings we also confirm that nonlinearity is required for predicting a nonlinear system, as demonstrated by the black column for $\eta_\textrm{max} = 1$ in Fig. \ref{fig:minimal-rc-lorenz-big-grid}.
However, \textit{how much} nonlinearity is needed?

\begin{figure}
\centering
\includegraphics[width=\columnwidth]{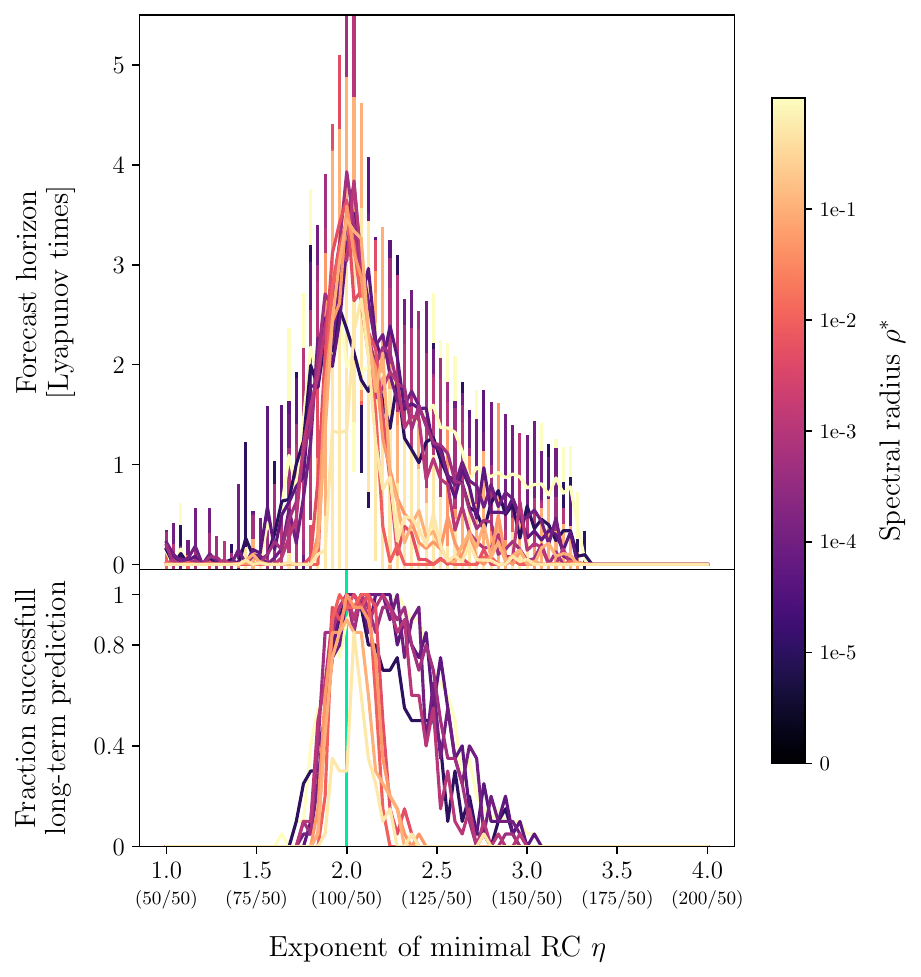}
\caption{\label{fig:lorenz-sweep}The short and long-term performance of minimal RCs reproducing the Lorenz system is presented.
We performed twenty experiments per parameter combination, with the upper plot reporting the mean and standard deviation of those twenty runs, while the lower one shows the fraction of successful reproductions.
The green line shows the true nonlinearity of the Lorenz system.
We show the results of $18\,240$ experiments.}
\end{figure}

We analyze the transition from the first column of failing predictions ($\eta_\textrm{max} = 1$) to the second column of successful predictions ($\eta_\textrm{max} = 2$) in more detail.
For that we apply the new fractional reservoir states for minimal RCs, meaning that the generalized reservoir states only contain two components: the linear one $\underline{r}$ and a single nonlinear one $\sqrt[\leftroot{1}\uproot{1}d]{\underline{r}^{\,n}}$.
We sweep the nonlinearity of the minimal RC from $\eta=1$ to $\eta=4$ with a denominator of $d=50$ in steps of two.
We use a block size of $b = 3$, a regularization parameter of $\beta=10^{-6}$, and iterate through each spectral radius from $10^{-5}$ up to $0.5$ including $0$.
For each combination of RC exponent $\eta$ and spectral radius $\rho^*$ we perform twenty realizations.
We train each minimal RC on $1\,000$ points and synchronize using $100$ points.
The results are presented in Fig. \ref{fig:lorenz-sweep}, where we analyze the short and long-term prediction.
The short-term prediction is measured using the forecast horizon, while we define a long-term prediction as successful, if the reconstructed Lyapunov exponent and the reconstructed correlation dimension do not differ more than $0.1$ from the original value.

For the short-term prediction we clearly observe a peak, when the nonlinearity in the data corresponds to the nonlinearity of the minimal RC.
We find this result to be stable for multiple hyperparameters.
Interestingly, this strong connection does not hold for the long-term prediction, where a reliable reconstruction of the attractor is possible even if the nonlinearity of the minimal RC exceeds the nonlinearity of the data.
Here we discover the relationship to be dependent on the spectral radius and observe that, in general, a lower spectral radius allows for a bigger deviation of the exponent of minimal RC against the exponent observed in data.
This implies that for an optimal short-term prediction the exponent of the minimal RC needs to exactly match the exponent of the data, while for a reliable long-term prediction a certain, small overestimation of the exponent in the data is allowed.

So far, we have only studied the Lorenz system, which contains a single order, integer nonlinearity of two.
However, we aim to explore how general our results are, which is why we expand this study on the fractional Halvorsen system with a controllable and non-integer nonlinearity in the following section.

\subsection{\label{sec:halvorsen-equal-xi}Fractional Halvorsen with $\xi_1 = \xi_2 = \xi_3$}
In this section we want to expand on the results of the Lorenz system and study whether the peak at the exponent in the data is due to an oddity of the Lorenz system and the power two, or whether we can observe a more general pattern.
Controlling the total nonlinearity in the Lorenz system is a nontrivial task due to the asymmetric equations and mixed nonlinearity, which is why we use the fractional Halvorsen system in Eqs. \ref{equ:halvorsen-equation}.

\begin{figure}
\centering
\includegraphics[width=\columnwidth]{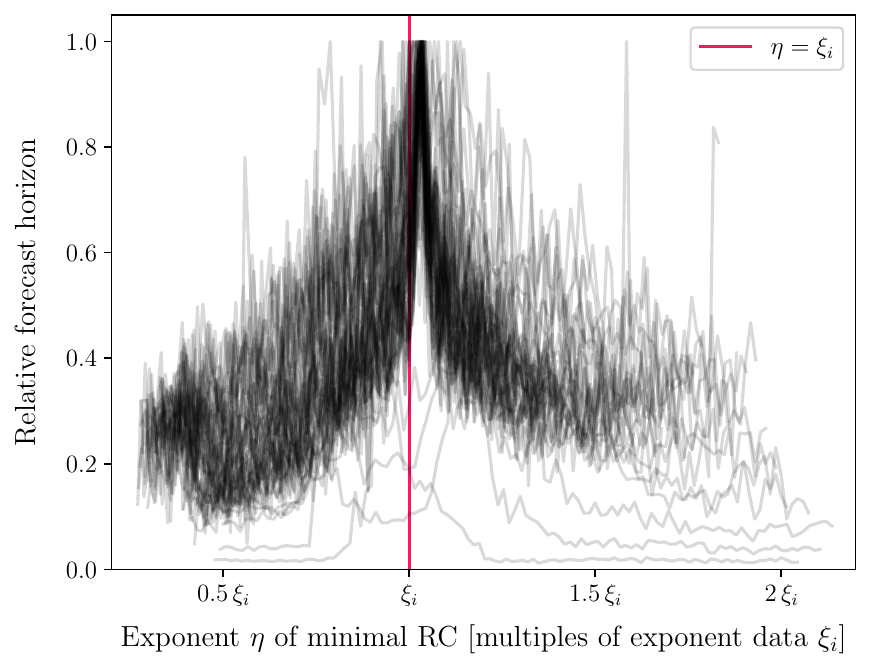}
\caption{\label{fig:halvorsen-equal-xi-lines}The mean relative forecast horizon for different nonlinear exponents $\eta$ of minimal RCs predicting the fractional Halvorsen system is shown.
Each gray line represents the mean for a different $\xi_i$.
For each parameter and exponent we perform seven runs and calculate the mean forecast horizon, which we normalize against the peak value.
However, we omit the error bars in the interest of readability.
We show the results of $62\,130$ experiments.}
\end{figure}

\begin{figure}
\centering
\includegraphics[width=\columnwidth]{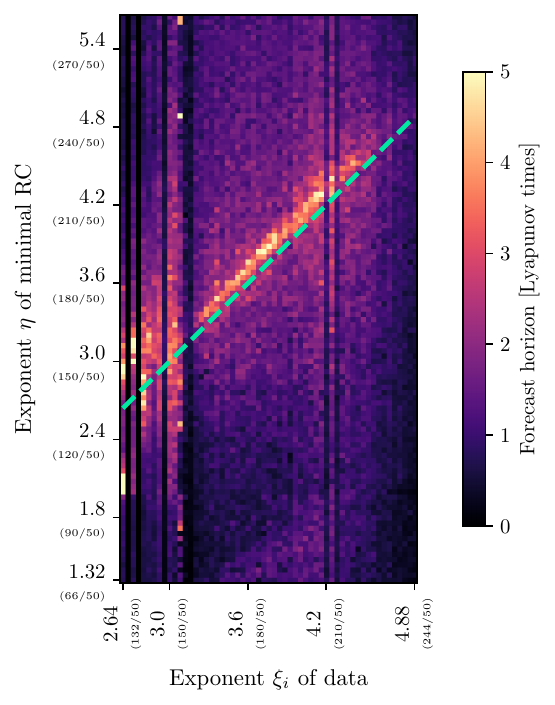}
\caption{\label{fig:halvorsen-heat}This plot shows the absolute forecast horizon for different exponents $\xi_i$ in data and $\eta$ in the model.
It uses the same underlying data of the fractional Halvorsen system as setup in Fig. \ref{fig:halvorsen-equal-xi-lines}.
Each tile shows the mean forecast horizon of five realizations and the green line indicates $\eta = \xi_i$.}
\end{figure}

For this experiment we set all exponents of the fractional Halvorsen system to the same value of $\xi_i = \xi_1 = \xi_2 = \xi_3$ and test numerators ranging from $n_i = 132$ up to and including $n_i = 280$ in steps of two with a denominator of $d = 50$ using a parameter of $a = 3.98$.
This corresponds to exponents ranging from $\xi_i = 2.64$ to $\xi_i=5.6$.
The exponents for the minimal RCs are ranging from $\eta=1.32$ to $\xi_i=5.6$ with the same denominator and step size.
We use a block size of $b=3$, a spectral radius of $\rho^* = 10^{-3}$, and a regularization parameter of $\beta = 10^{-6}$.
We train each minimal RC on $5\,000$ points, synchronize using $1\,000$ points, and perform seven runs per parameter combination.
The findings for this experiment are shown in Figs. \ref{fig:halvorsen-equal-xi-lines} \& \ref{fig:halvorsen-heat}.

We observe a clear peak at $\eta = \xi_i$, indicating that hitting the exact nonlinearity of the data is important for a successful prediction.
Interestingly, we note that even overshooting the nonlinearity in the data will not generally improve the predictive power.
The single lines in Fig. \ref{fig:halvorsen-equal-xi-lines} increasing after $1.5\,\xi_i$ are systems with a very low Lyapunov exponent and thus easier to predict.
The bulk of interesting results lies before and around $\eta = \xi_i$.
Additionally, we want to note the width of the peak in Fig. \ref{fig:halvorsen-equal-xi-lines}, which seems to be (roughly) constant using a relative $x$-axis.
This finding could be useful when building a non-integer library in Sec. \ref{sec:smart-library}, indicating that, for larger exponents, a less precise guess in absolute terms is required for a reasonable prediction performance than for smaller ones.

\subsection{Fractional Halvorsen with $\xi_1 = \xi_2 \neq \xi_3$}
So far, we have only studied the fractional Halvorsen system with all equal exponents.
Due to the symmetry of the equations and the equality of the exponents, the data only contained a single nonlinearity.
Here, we want to systematically study the inclusion of two different exponents.

For this case we set $\xi_1 = \xi_2 = \xi_{1,2}$ in Eqs. \ref{equ:halvorsen-equation} and set it to differ from $\xi_3$.
For $\xi_{1,2}$, $\xi_{3}$, and $\eta$ we use numerators ranging from $n = 54$ up to $n = 280$ in steps of two with a denominator of $d=50$.
This corresponds to values from $\eta_{\textrm{min}}=1.08$ to $\eta_{\textrm{max}}=5.6$ for all exponents.
We use a parameter of $a=3.98$.
For each set of parameters, we perform five experiments.

\begin{figure}
\centering
\includegraphics[width=\columnwidth]{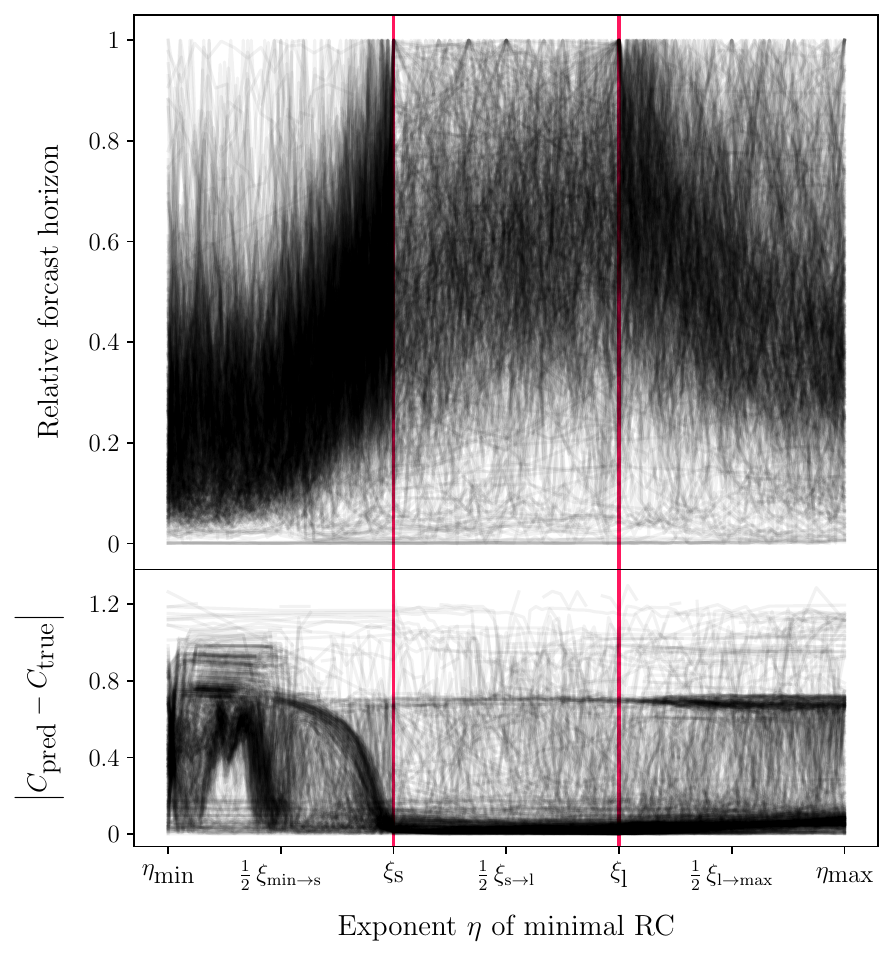}
\caption{\label{fig:halvorsen-two-exponents}The prediction performance for predicting the fractional Halvorsen system with two equal, fractional exponents is shown: $\xi_1=\xi_2\neq\xi_3$. We show the mean relative forecast horizon for different nonlinear exponents of minimal RCs in relative terms in upper plot and the correlation dimension error in the lower plot.
We build this figure from $56\,031$ experiments resulting in $983$ trajectories.}
\end{figure}

We show the results in Fig. \ref{fig:halvorsen-two-exponents}.
We order the values such that $\xi_\textrm{s} =  \min{\left( \xi_{1,2},\;\xi_3 \right)}$ and correspondingly $\xi_\textrm{l} =  \max{\left( \xi_{1,2},\;\xi_3 \right)}$.
We do this simplification since we find that the results do not change depending on the ordering of the two exponents.
For the sake of readability, we define the relative distance between them as
\begin{equation}
\label{equ:notation-xi-ab}
    p\,\xi_{a\to b} \coloneq \xi_a + p\,\left(\xi_b - \xi_a\right)\;\;\;.
\end{equation}

While we cannot find a quantitative pattern in Fig. \ref{fig:halvorsen-two-exponents}, we can make qualitative statements about this experiment.
Looking at the interval $\left[\eta_{\textrm{min}},\;\xi_\textrm{s}\right]$ we find the previous pattern of the prediction performance increasing until we hit the first nonlinearity of the data.
In the interval $\left[\xi_{\textrm{s}},\;\xi_\textrm{l}\right]$ between the two nonlinearities of the data, there is neither a clear pattern nor consistent peaks.
The last interval $\left[\xi_{\textrm{l}},\;\eta_{\textrm{max}}\right]$ shows again a familiar pattern of the performance decreasing when the exponent of the minimal RC gets larger than the exponent present in the data.
The interesting finding of this experiment is the peaking of the prediction performance at $\xi_\textrm{s}$ and $\xi_\textrm{l}$:
The prediction performance is best when one of the nonlinearities is hit.
Unexpectedly, the smallest one seems to be the most important one providing the most contribution to a successful prediction.

This indicates that for mixed nonlinearities we do not observe a significant improvement after including the smallest nonlinearity.
It seems that including the smallest nonlinearity is more important than finding all of them.
This is an important finding, since real-life data cannot be expected to contain only a single nonlinearity, and this finding hints that finding the smallest one is the most valuable one regarding the prediction performance.

\subsection{\label{sec:three-etas}Fractional Halvorsen with $\xi_1 \neq \xi_2 \neq \xi_3$}
In this section, we want to complete the analysis of the fractional Halvorsen system by studying the case where all three exponents differ from each other.
A coordinated study, as performed in previous sections, is not feasible due to the huge number of possible combinations.
Instead, we randomly pick three numerators from $52$ to $280$ with the common denominator of $50$ for the exponents $\xi_i$, and, if the trajectory is a valid chaotic system, we sweep the nonlinearity of the minimal RC from $\eta_\textrm{min}=\nicefrac{52}{50}$ to $\eta_\textrm{max}=\nicefrac{280}{50}$ in steps of two in the numerator.
The exponents are ordered by magnitude and named $\xi_\textrm{s} < \xi_\textrm{m} < \xi_\textrm{l}$.
We want to test whether our qualitative result of the previous section also holds for the case of three different exponents.

\begin{figure}
\centering
\includegraphics[width=\columnwidth]{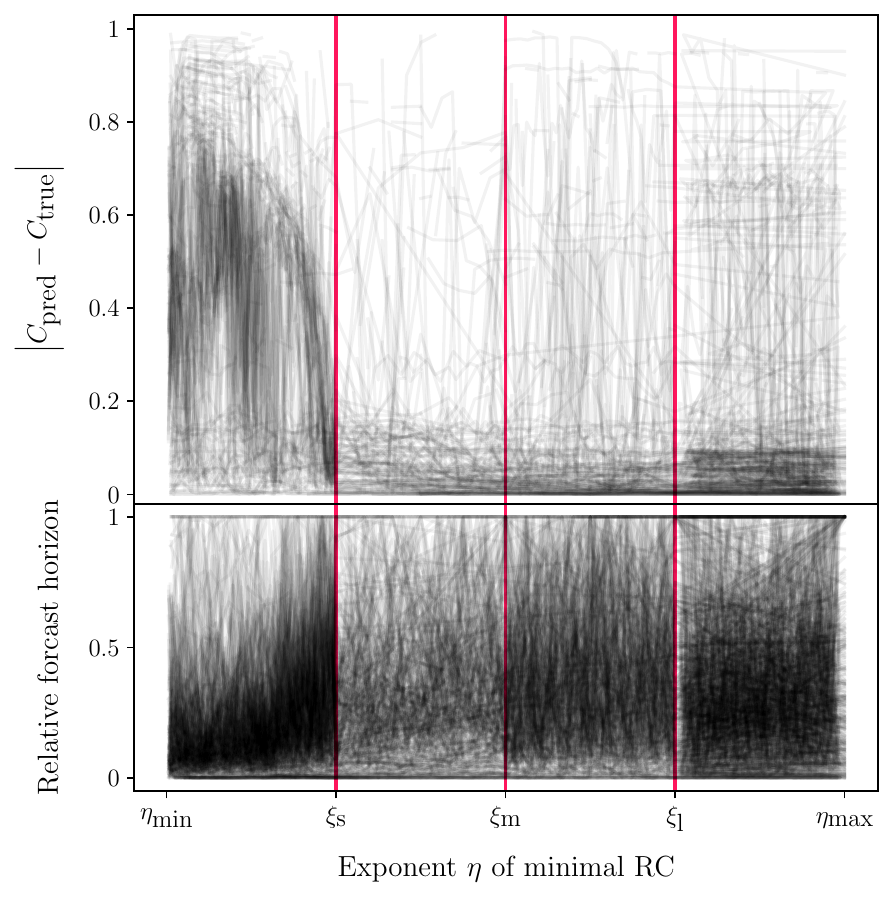}
\caption{\label{fig:halvorsen-three-exponents}The prediction performance for predicting the fractional Halvorsen system with three different fractional exponents is shown: $\xi_1\neq\xi_2\neq\xi_3$. We show the absolute difference between the true correlation dimension and the predicted correlation dimension in upper plot. The lower plot shows the relative forecast horizon for different nonlinear exponents of minimal RCs in relative terms.
We build the $564$ trajectories from $32\,148$ experiments.}
\end{figure}

The results are presented in Fig. \ref{fig:halvorsen-three-exponents}, where we can confirm the pattern of Fig. \ref{fig:halvorsen-two-exponents}, in which the error of the predicted correlation dimension drops to zero as soon as the nonlinearity of our estimator exceeds the smallest nonlinearity of the data.
Due to the small number of trajectories, other patterns are not qualitatively testable in Fig. \ref{fig:halvorsen-three-exponents}.
Nevertheless, confirming the special situation of the smallest nonlinearity present in data is an important finding.

\section{\label{sec:applications}Applications}
From our findings, we identify two possible applications.
The first application consists of determining the smallest nonlinearity present in the data, by sweeping the nonlinearity in the minimal RC model and observing its output.
For the second application we want to take the findings from minimal RC and apply them to traditional RC architectures by extending traditional RCs with fractional nonlinearities.

\subsection{\label{sec:finding-smallest-nonlinearity}Smallest nonlinearity in data}
A possible application of this framework consists of discovering the smallest nonlinearity present in data.
We have shown previously that the correlation dimension of the predicted values approaches the correlation dimension of observed data for the first time if the exponent of the minimal RC matches the smallest exponent present in the data.
We can use this observation to build a test for the smallest exponent present in data.

For that, we first determine the correlation dimension of the time series.
We then sweep through a range of non-integer exponents for the minimal RC.
For each exponent we calculate the correlation dimension of the predicted time series.
We repeat the same for surrogate versions of the time series, in order to assure that the observed effect really stems from the nonlinearity of the data.
When the predicted correlation dimension matches the true one and is outside the surrogates ones, we found an approximation for the smallest nonlinearity present in data.
If the measure of the time series does not exceed the surrogate measure, the determination of a smallest nonlinearity failed with this method.

\begin{figure*}
\centering
\includegraphics[width=\textwidth]{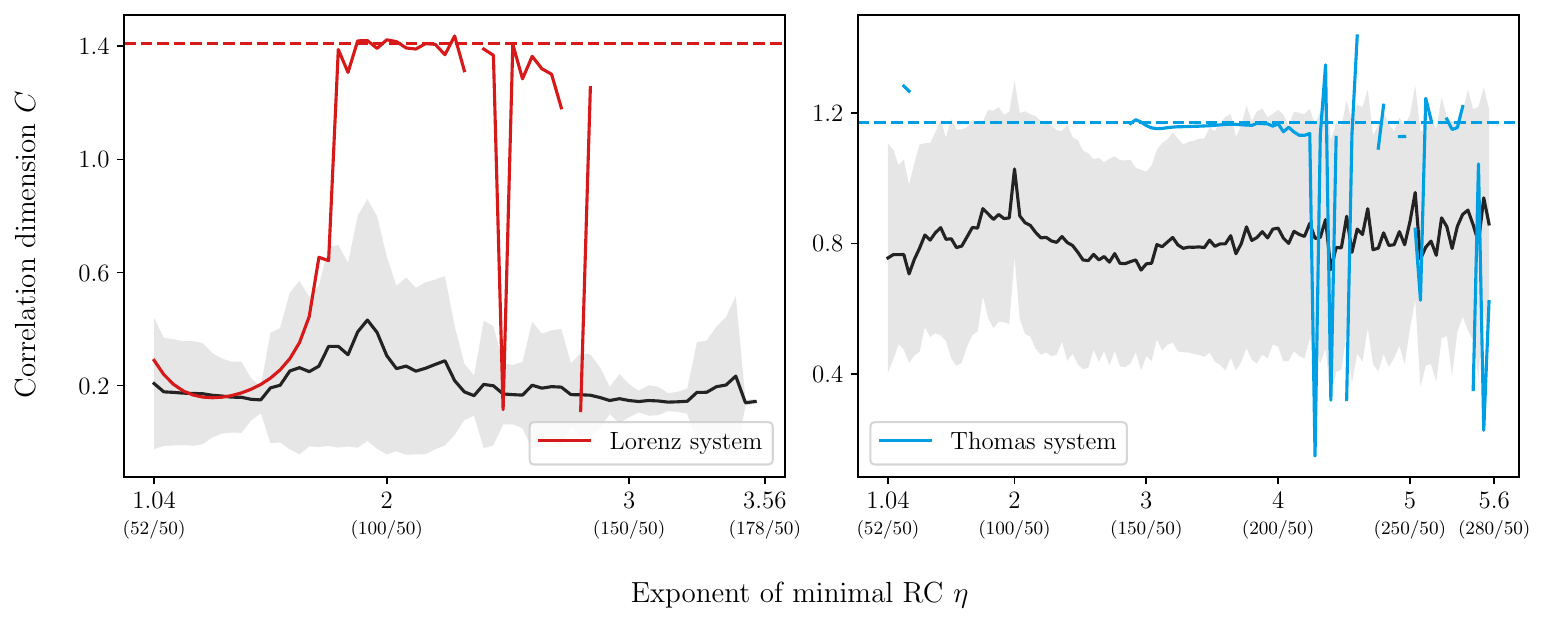}
\caption{\label{fig:minimal-nonlinearity}The correlation dimension of the predictions using the reduced minimal RC with one linearity against the surrogate background for the Lorenz and Thomas system is shown. The black line and gray area represent the mean and one standard deviation of the correlation dimension when training the reduced minimal RC with FT surrogates. The dashed colored line shows the real correlation dimension of the data. Each plot is the result of $6\,375$ optimizations.}
\end{figure*}

We test our method on some chaotic systems with known nonlinearity, some real-world data with unknown nonlinearity, and present the results in Tab. \ref{tab:reconstructed-nonlinearities} and Figs. \ref{fig:minimal-nonlinearity}f.
In Sec. \ref{sec:benchmarking-rc-sindy} we then benchmark our results against classical RCs and SINDy.

The chaotic systems of choice are the Lorenz system, the classical Halvorsen system with $\xi_i=2$, and the Thomas system.
We parameterize our minimal RCs with a block size of $b=3$, a target spectral radius of $\rho^*=0.1$, and a regularization parameter of $\beta=10^{-6}$.
We synchronize our model using $100$ data points and train it on $1\,000$ data points.
The nonlinearity exponent $\eta$ is swept from $\nicefrac{52}{50}$ to $\nicefrac{280}{50}$ in steps of two in the numerator.
Fig. \ref{fig:minimal-nonlinearity} shows the result for the Lorenz system and the Thomas system.

For the Lorenz system we observe the predicted correlation dimension to rise very quickly when the nonlinearity in the minimal RC model approaches two, the nonlinearity in the data.
The value for the correlation dimension differs significantly from the surrogate background rendering our estimate of it confident.
Interestingly, for the Thomas system the correlation dimension becomes a computable number, and instantly reaches the correlation dimension of the data for an exponent of $2.92$, which is very close to $3$.
The Thomas system has its nonlinearity in the sine function, and an exponent of $3$ constitutes the first nonlinear term of its Taylor approximation.
We note that the predicted correlation does not stand out strongly from the linear surrogate background, as it does for the Lorenz system.
A possible explanation for this is that the Thomas system contains only a small degree of nonlinearity.
This can be seen in its largest Lyapunov exponent of $\lambda \approx 0.01$, which is barely positive (compared to $\lambda \approx 0.9$ for the Lorenz system).
Additionally, the terms in the Taylor approximation of the sine function scale with the factorial of the exponent, keeping the nonlinear effect apparently small.
Nevertheless, it is interesting that the observed smallest nonlinearity seems to correspond to the first nonlinear term of the Taylor expansion.
We have seen that for synthetic systems we are able to coarsely determine the smallest nonlinearity present in data making us confident to apply this method to real world data.

For the financial data we perform our test on three different stock indices: the MSCI World Index tracking publicly traded large- and mid-cap companies across the developed world; the S\&P 500 Index tracking the 500 largest, publicly traded, U.S. companies; and the STOXX Europe 600 Index covering 600 publicly traded, European companies spanning from small- to large-cap.
For each index we use the daily closing value starting from 1\textsuperscript{st} March 2005 up to 31\textsuperscript{st} January 2025 for calculating the daily return (percentage change).
This results in roughly $5\,000$ data points for each index.
We use the daily returns, since there is evidence that they are (weakly) stationary\cite{Cont2001} making it appropriate for our setup.
For each index we use the same parameterization for the minimal RCs consisting of a block size of $b = 5$, a spectral radius of $\rho^* = 0.99$, and a regularization parameter of $\beta=10^{-6}$, and we use the first $500$ steps as synchronization steps.

\begin{figure}
\centering
\includegraphics[width=\columnwidth]{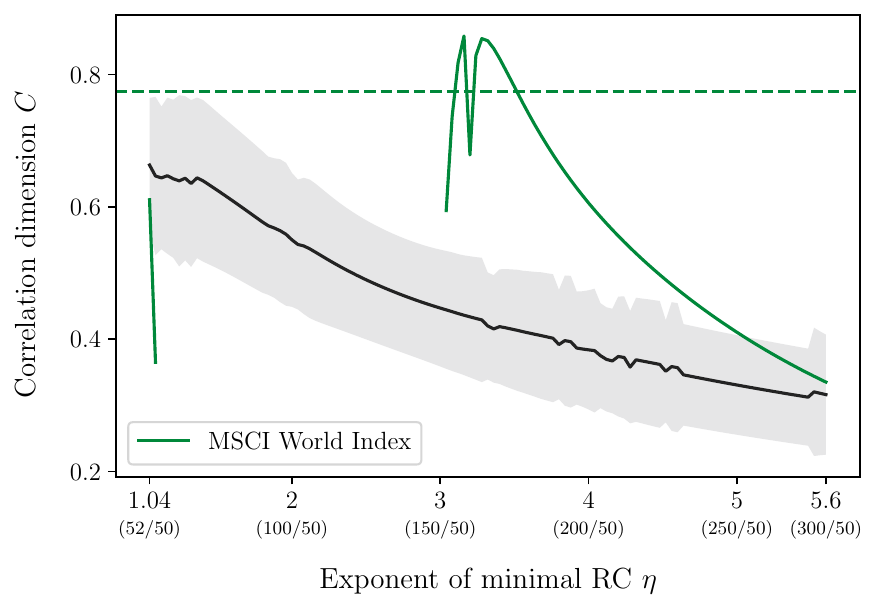}
\caption{\label{fig:nonlinear-msci}The correlation dimension of the predictions of the MSCI World Index data using the reduced minimal RC model is shown. The black line with the gray area shows the mean and one standard deviation of the surrogate background. The dashed line represents the real correlation dimension of the MSCI World Index returns. This figure is a result of $6\,375$ experiments.}
\end{figure}

The results are presented in Tab. \ref{tab:reconstructed-nonlinearities}, where we can see that the test was successful for each index and we are able to determine a minimal nonlinearity for each.
Here, we note that predicting the stock indices was obviously unsuccessful.
However, it seems that the unsuccessful prediction was enough to capture the nonlinearity in the data, as we observe a peak like in Fig. \ref{fig:nonlinear-msci} for every index.

We find it difficult to put our findings into context, as comparable studies exploring the smallest nonlinearity in financial data have not yet been conducted.
Therefore, our results should be viewed as an exploratory proof-of-concept demonstration rather than conclusive evidence for existence of underlying nonlinearities in financial data.
Given the well-known stochasticity of financial data\cite{Fama1965} any attempt to infer deterministic structure must be approached with caution, as nonlinear effects identified by our method could also be influenced by transient effects (\textit{e.g.}, market reactions to interest rate cuts) or structural breaks (\textit{e.g.}, sudden regime shifts during crises).
Such events may lead to an overestimation of deterministic structure.
Nevertheless, we find the observed patterns compelling, and believe they warrant continued investigation into the role of nonlinearity in financial markets.

Additionally, we note at this point that this proposed method is not guaranteed to work for arbitrary datasets.
It can happen, as it did in our case for the Atlantic meridional overturning circulation (AMOC) dataset\cite{AMOC2023}, that the reconstruction fails.
The reconstructed correlation dimension does not exceed the surrogate background and does not meet the correlation dimension measured in the dataset.
However, this does \textit{not} imply that the dataset does not contain any nonlinearity.
Instead, it simply means that our method is not able to detect any nonlinearity. In the case of the AMOC data it may simply be due to the fact that not enough training data was available as the AMOC time series is considerably shorter than that of the financial data.
Further, in-depth analyses are required to identify the necessary prerequisites for the method to work for real data. 
Nevertheless, the results on financial data already point to the fascinating possibility of being able to derive the degree of nonlinearity of the underlying process for a real dataset.

To estimate the smallest nonlinearity present in a dataset, we recommend the following procedure to practitioners: First, identify a suitable parametrization of the full minimal reservoir computer that reliably models the system, using techniques like Bayesian optimization if needed.
Next, define a list of candidate fractional nonlinearities with a resolution adapted to the available computational resources.
Finally, apply the reduced minimal RC framework with a single nonlinearity across this list to both the original data and its surrogate versions.
The first performance peak, where the model outperforms surrogates, can then be interpreted as an estimate of the smallest nonlinearity in the data.

\subsubsection{\label{sec:benchmarking-rc-sindy}Results of comparable methods}
Minimal RCs, as a subset of classical RCs, are conceptually very similar to SINDy models, as both use a library to construct nonlinear features from data.
For this reason, we benchmark our results against classical RCs and SINDy.

\paragraph{Classical reservoir computers}
For classical RCs, we use the same approach as for minimal RCs, employing fractionally exponentiated, generalized reservoir states. We use two different reservoir sizes in our experiments: a small network with $d=100$ nodes and a large network with $d=1\,100$ nodes.
For each setup we use random networks created by the Erdős--Rényi algorithm\cite{erdosrenyi} and use a target spectral radius of $\rho^* = 0.2$.
We synchronize the reservoir to the data using $1\,000$ data points and train on $4\,000$ steps performing a ridge regression with a regularization parameter of $\beta=10^{-4}$.
We use fractions ranging from $\nicefrac{4}{50}$ to $\nicefrac{400}{50}$ in steps of $\nicefrac{4}{50}$ as fractional exponents for the additional nonlinearity.

Using this setup, we were not able to reconstruct the smallest nonlinearity of the Lorenz system. 
In line with the results from \textcite{Herteux2020} and our later results in Sec. \ref{sec:smart-library}, we note an improved performance when introducing generalized reservoir states.
However, the increase is independent of the amount of nonlinearity making it impossible to reconstruct the underlying nonlinearity.

\paragraph{SINDy}
We use our fractional library to study whether we can reconstruct the smallest nonlinearity using SINDy.
We use the global nonlinearities $\mathbf{H} = \left\{\nicefrac{10}{50},\,\nicefrac{20}{50},\,\dots,\,\nicefrac{300}{50}\right\}$ and use each $\eta \in \mathbf{H}$ to construct a nonlinear library with a single nonlinearity.
Due to the comparatively large library, we employ a relatively large parameter value of $\alpha=10$ to enforce sparsity.
We train each SINDy model on $5\,000$ time steps.

The results are shown in Tab. \ref{tab:reconstructed-nonlinearities}, where we can see that the reconstruction fails for the real-world datasets.
However, for the synthetic systems we find rather interesting results.
For the Lorenz system we find that a nonlinearity of $\nicefrac{60}{50}$ is sufficient to reproduce a slightly `wobblier' version of the attractor which, nevertheless, has the correct correlation dimension.
As in our previous work\cite{Prosperino2025}, we again find different sets of differential equations that produce a chaotic, butterfly-shaped attractor---this time even with different nonlinear terms as compared with the original set of the Lorenz equations.
They are obviously not the only ones that produce this very butterfly-shaped attractor.

\begin{figure}
\centering
\includegraphics[width=\columnwidth]{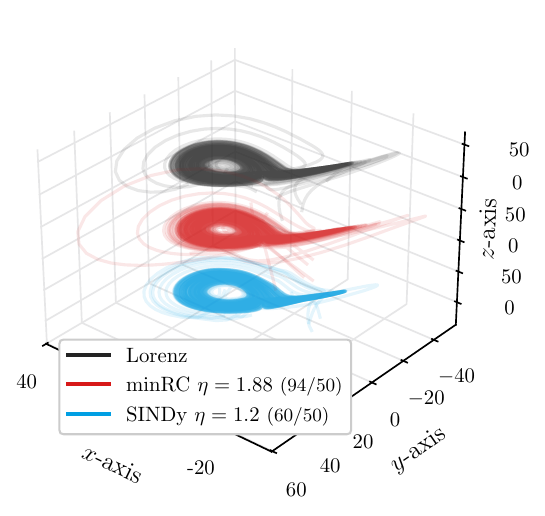}
\caption{\label{fig:all-lorenz}We show the classical Lorenz attractor on top in black.
For minimal RC and SINDy we show the reconstructed Lorenz attractor with the minimal working nonlinearity, namely a nonlinearity of $\eta=1.88$ for minimal RC in the middle in red and a nonlinearity of $1.2$ for SINDy on the bottom in blue.
The setup of the models is as described in the text. For visualization purposes we separate the attractors in the phase space across the $z$-axis. We show twenty trajectories per model, each with a random initial condition consisting of a random integer between $-20$ and $20$ in each coordinate.}
\end{figure}

In Fig. \ref{fig:all-lorenz} we show a comparison of the three different attractors: the real one, the one reconstructed with minimal RC, and lastly the one reconstructed with SINDy.
All three show the same butterfly-like shape and are visually virtually indistinguishable.
However, they all rely on vastly different amounts of nonlinearity.

It seems that SINDy is more stable than minimal RC and requires way less nonlinearity than present in data to successfully approximate a trajectory.
While an interesting result, especially the alternative descriptions of the chaotic systems, we find it to be not suited for discovering the smallest `true' nonlinearity present in data.
\begin{table}
\caption{\label{tab:reconstructed-nonlinearities}In the first part of this table we show the real and predicted smallest nonlinearity $\mu$ for traditional chaotic systems.
In the second part we present the reconstructed smallest nonlinearity of financial systems, and lastly we indicate a failed reconstruction of climate data.}
\begin{ruledtabular}
\begin{tabular}{lccc}
System&$\mu_\textrm{real}$&$\mu_\textrm{minRC}$&$\mu_\textrm{SINDy}$\\
\hline
Lorenz&2&1.88&1.2\\
Classical Halvorsen&2&1.96&failed\\
Thomas&$3$\footnotemark[1]&2.92&1.8\\
\hline
MSCI World Index&n.a.& $3.12$&failed\\
S\&P 500 Index&n.a.& $1.64$&failed\\
STOXX Europe 600 Index&n.a.& $5.32$&failed\\
\hline
AMOC&n.a.&failed&failed
\end{tabular}
\end{ruledtabular}
\footnotetext[1]{The Thomas system has a sine nonlinearity. Here we consider the first nonlinear term of the Taylor expansion of the sine function, $\nicefrac{x^3}{3!}$, as smallest nonlinearity.}
\end{table}

\subsection{\label{sec:smart-library}Smart non-integer library}
\begin{figure}
\centering
\includegraphics[width=\columnwidth]{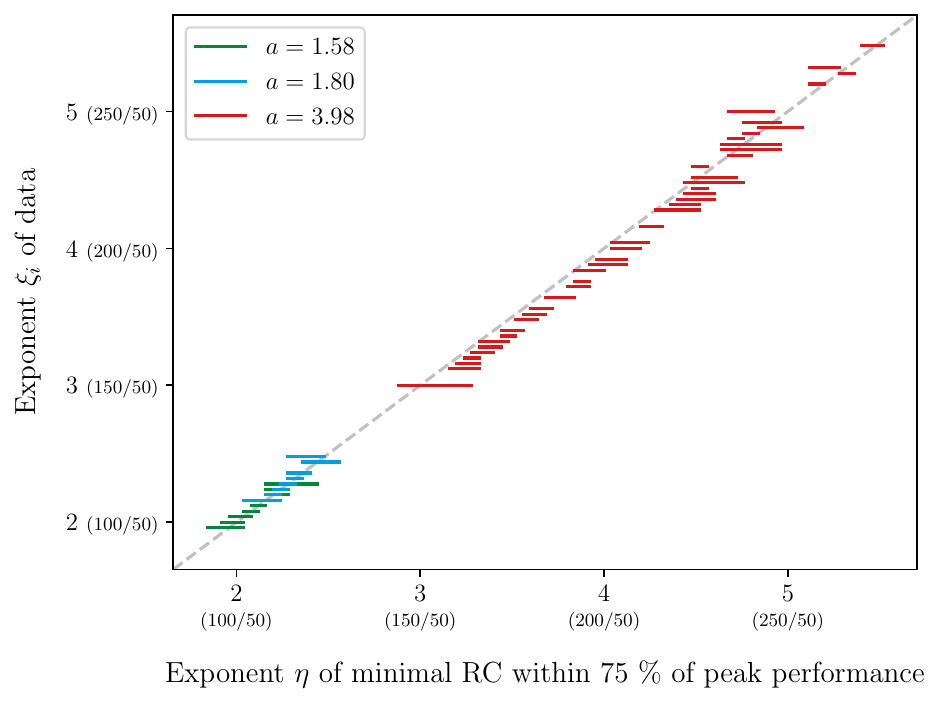}
\caption{\label{fig:fwhm}For each exponent $\xi_i$ of the fractional Halvorsen system, we show the range of exponents for minimal RCs, where the relative forecast horizon lies between $75\%$ of their maximal value.
For this plot we compiled three different parameters $a$ of the fractional Halvorsen system.
The gray line indicates the identity where $\eta = \xi_i$.
We compile the results of in total $86\,070$ separate experiments.}
\end{figure}

For the second application we want to transfer our findings discovered for minimal RCs to the traditional RC architecture.
While generalizing reservoir states is not a new idea\cite{Herteux2020, Ohkubo2024}, we propose extending this idea with fractional powers.
We have seen that minimal RCs work best, when the generalized reservoir state contains the nonlinearity present in the data.
For this reason, we propose including fractional powers of the reservoir states in minimal RC.
To determine the number of powers to include, we revert to our results from Sec. \ref{sec:halvorsen-equal-xi} in Fig. \ref{fig:halvorsen-equal-xi-lines}:
inspired by the full width at half maximum metric commonly used in optics, we calculate the full width at $75\%$ from the peak performance in Fig. \ref{fig:halvorsen-equal-xi-lines}.
While we only used a parameter of $a=3.98$ for Fig. \ref{fig:halvorsen-equal-xi-lines}, we extend the study for Fig. \ref{fig:fwhm} to include parameters of $a=1.58$ and $a=1.80$, in order to get a feeling for the width at lower powers.
The results are shown in Fig. \ref{fig:fwhm}, where we see the width for $75\%$ of the peak performance staying constant across all exponents and different parameters, indicating faintly that this result may be generalizable.

With this finding we can choose fractional exponents whose width will cover the whole space between two integers.
One possible realization of these exponents can be constructed with
\begin{equation}
    \label{equ:spacing}
    \underline{\tilde{r}}_{[1,\;2]} = \begin{pmatrix}
    \underline{r}^1 & \underline{r}^{\tfrac{54}{50}} & \underline{r}^{\tfrac{66}{50}} & \underline{r}^{\tfrac{78}{50}} & \underline{r}^{\tfrac{90}{50}} & \underline{r}^2
    \end{pmatrix}^\textrm{T}\;\;\;.
\end{equation}
Here, the subscript $[1,\;2]$ indicates that the fractions simply represent the spacing between the integer powers $1$ and $2$, and this idea can be generalized up to an arbitrary integer power.
In later applications we extend them to span up to an integer power of $3$.
We find our results to be robust against the exact choice of fractional powers.
While we acknowledge this being a rather rudimentary approach with room to improvement for finding the optimal fractional powers, we find this approach sufficient for the scope of this work.

We want to test whether the findings for minimal RCs can be transferred to work on traditional RCs.
We perform this test by predicting the Lorenz system using three different architectures:
firstly, we use a RC with a dimensionality of $d=100$.
The new architecture also uses a dimensionality of $d=100$ but generalizes the reservoir states $\underline{r}$ to $\underline{\tilde{r}}$ by including fractional powers of the reservoir state up to a power of $3$ with the spacing of Eq. \ref{equ:spacing}.
However, since including these additional powers increases the size of the output matrix by a factor of 10, we need to test our proposed change against a reservoir resulting in the same sized output matrix.
This results in a third RC model of with a dimensionality of $d=1\,100$.
For all three models we use the same hyperparameters: a spectral radius of $\rho^*=0.2$ on a random network and a regularization parameter of $\beta=10^{-4}$.
We train all models on $4\,000$ data points and use $1\,000$ data points for the synchronization phase.

\begin{figure}
\centering
\includegraphics[width=\columnwidth]{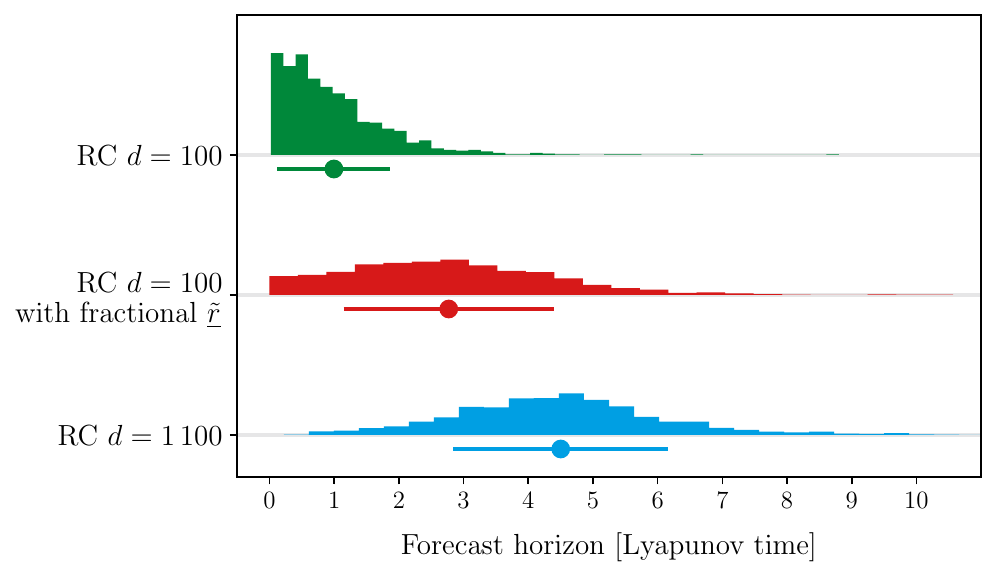}
\caption{\label{fig:library}We show the prediction performance for three different RCs: Green represents the small reservoir, red the small reservoir with fractional reservoir states, and blue the large reservoir.
For each RC we performed $1\,000$ experiments and show the distribution of the forecast horizons, and the mean with one standard deviation below.}
\end{figure}

While we see in Fig. \ref{fig:library} that the small reservoir with fractional reservoir states performs worse than the large reservoir, it easily outperforms the standard small reservoir.
Therefore, we can improve the performance of small reservoirs by generalizing their reservoir states to include fractional powers.
This approach can be used to enhance the performance of physical RC implementations in situations where increasing the reservoir size is not feasible due to constraints in the hardware fabrication process, such as the limited number of neurons available on neuromorphic chips or increasing production costs associated with larger physical reservoirs\cite{Stepney2024}.

For practitioners, we note that the exact choice of fractional exponents is not critical.
Instead, what matters most for small reservoirs is the presence of fractional nonlinearities, not their precise values.
The proposed spacing in Eq. \ref{equ:spacing} offers a practical template that balances expressiveness and complexity.
Importantly, the number of fractional terms included should be adapted to the size of the available training data, as too many features can lead to overfitting.
This fractional augmentation offers a computationally efficient way to enhance prediction quality in resource-constrained settings without increasing reservoir size.

\section{\label{sec:discussion}Conclusion}
In this work we systematically explored the relationship between the nonlinearities in input data and those introduced in RC models.
Building on the minimal RC framework, we proposed a tailored architecture with a single, tunable nonlinearity parameter, allowing us to isolate and precisely control the degree of nonlinearity in the model.
Using a novel fractional Halvorsen system, we generated chaotic datasets with adjustable nonlinear structure and evaluated the prediction performance across a wide range of reservoir states nonlinearities.

While we restricted ourselves to even numerators to avoid imaginary-valued reservoir states, the idea of studying imaginary reservoir states presents a compelling direction for future research.

Through extensive experiments, we found that short-term forecasting performance is maximized when the nonlinearity in the minimal RC model matches the nonlinearity of the input data.
In other words, the best predictions occur at a tuned `nonlinearity match' between data and model, whereas mismatches, by using a model that is too linear or too nonlinear relative to the data, consistently degrade performance.
This confirms our core hypothesis and directly demonstrates that optimally tailored reservoir states yield superior results.

Our work strongly corroborates the `Catch-22'\footnote{To fully appreciate the depth of this term, we wholeheartedly recommend the novel ``Catch-22'' by Joseph Heller.} described by \textcite{Zhang2023}.
Specifically, their results show that NGRCs perform well only when the exact governing terms of the system are known; but if those are known, there is no need for a machine learning model.
We observe the same pattern using minimal RCs.
The best performance is achieved when the nonlinearity of the model matches the nonlinearity of the data.
This represents a limitation of such models and emphasizes the importance of developing reliable methods for inferring the underlying system structure directly from data.
Conversely, these results support the idea that the success of classical RCs lies in their ample spectrum of nonlinearities.

However, importantly, we also observed that for systems with multiple nonlinearities, it is often the smallest nonlinearity in the data that dominates prediction performance.
This insight enables us to use our framework in reverse: By sweeping through model nonlinearities and observing the resulting performance, we were able to estimate the minimal nonlinearity present in a given time series.
While we were not able to observe effectiveness across all datasets studied, applying this method to both synthetic and real-world financial data demonstrated its practical value in uncovering underlying nonlinear structures.

Interestingly, when applying SINDy with a restricted fractional library, we were able to identify governing equations that reproduce the characteristic butterfly-shaped Lorenz attractor using nonlinearities significantly smaller than two, suggesting that simpler functional forms can approximate complex dynamics under certain conditions.

Finally, we transferred our findings to classical RC architectures and demonstrated that incorporating fractional, generalized reservoir states leads to an improvement in predictive performance.
This has direct implications for physical RC platforms, where increasing the number of reservoir nodes may not be feasible due to hardware or economic constraints.
By enhancing the expressiveness of the reservoir through non-integer polynomial transformations—--rather than scaling the system size--—we enable a more compact yet powerful representation of the input dynamics.
This approach offers a structured way to increase the abilities of physical reservoirs without increasing their structural complexity, making it a viable strategy for high-performance prediction in embedded or resource-constrained environments.

Our work offers both a theoretical and practical step forward in understanding and designing reservoir computers that are better aligned with the complexity of the data they aim to model.

\begin{acknowledgments}
We would like to thank {Allianz} {Global} {Investors} for providing data and computational resources.
\end{acknowledgments}

\section*{Data Availability Statement}
The data that support the findings of this study are available from the corresponding author upon reasonable request.

\appendix

\section{\label{sec:metrics}Metrics}
Here, we want to define the methods used for quantifying the quality of a prediction.
As in similar work, we use the forecast horizon to quantify the short-term prediction power.
The so-called long-term `climate' of an attractor is measured by the largest Lyapunov exponent in combination with the correlation dimension.

\subsection{\label{sec:largest-lyapunov-exponent}Largest Lyapunov exponent}
The Lyapunov exponent formalizes the concept of sensitive dependence on initial conditions of chaotic systems.
Defining the distance $\delta$ between two nearby points $\underline{x}$ and $\underline{x} + \underline{\varepsilon}$, experimentally an exponential increase of this distance $\delta$ can be observed for chaotic systems modeled by
\begin{equation}
    \delta(t) = \delta(0)\,\exp{\lambda\,t}\;\;\;.
\end{equation}
Here, $\lambda$ describes the largest Lyapunov exponent and is a measure for how fast two nearby trajectories diverge.

We calculate the largest Lyapunov exponent from the data using the algorithm introduced by \textcite{Rosenstein1993}.
For this, we track the evolution of initially close points in phase space by identifying nearest neighbors in the time series and measuring the average logarithmic divergence over time.
The slope of this divergence, computed over a selected time interval yields an estimate of the maximal Lyapunov exponent.
To ensure valid comparisons, pairs of points are filtered to avoid temporal proximity and to allow sufficient forecast length.

Using the largest Lyapunov exponent $\lambda$, a Lyapunov time $\tau_\lambda \coloneq \lambda^{-1}$ can be defined, representing the characteristic time scale over which trajectories in phase space remain close.
This quantity acts as a natural reference time scale for analyzing and contrasting the dynamics of various systems.

\subsection{\label{sec:correlation-dimension}Correlation dimension}
The correlation dimension is a widely used measure to estimate the fractal dimensionality of strange attractors and provides insight into the geometric complexity of a system’s long-term behavior.
It is based on the idea of quantifying how the number of point pairs within a certain distance $r$ scales with $r$ itself.
The correlation sum $\mathcal{C}(r)$ is defined as the fraction of pairs whose mutual distance is smaller than $r$ by
\begin{equation}
    \mathcal{C}(r) = \lim_{N \to \infty} \frac{1}{N^2} \sum_{t_1\neq t_2} \Theta{\left( r - \lVert \, \underline{x}(t_1) - \underline{x}(t_2) \rVert \right)}\;\;\;.
\end{equation}
Here, $\Theta$ is the Heaviside step function returning one when the distance is smaller than $r$ and zero when the distance is larger than $r$.

Experimentally it has been discovered that for self-similar, strange attractors the power law
\begin{equation}
    \mathcal{C}(r) \sim r^C
\end{equation}
holds over a range of $r$.
To define a meaningful range of radii, we compute the lower and upper bounds as the 5\textsuperscript{th} and 95\textsuperscript{th} percentiles of all pairwise distances, respectively.
This ensures that the correlation sum is evaluated over a scale that captures the system's spatial structure.

The scaling factor of the power law $C$ is the correlation dimension and describes how densely the points fill space as the scale $r$ decreases.

We calculate the correlation dimension using the algorithm by \textcite{Grassberger1983} by embedding the time series in phase space and estimating how the number of point pairs within a radius $r$ scales with $r$.
To do this efficiently, we organize the data using a binary space-partitioning data structure that stores the points in a $k$-dimensional space\cite{Maneewongvatana2002}.
Using this structure, we compute the correlation sum $\mathcal{C}(r)$ over a range of radii $r$.
The slope of the double logarithmic graph of $\mathcal{C}(r)$ over $r$ yields the correlation dimension $C$.

\subsection{\label{sec:forecast-horizon}Forecast horizon}
For quantifying the short-term prediction, we use the forecast horizon as a measure that describes the time for which the error between the true trajectory $\underline{x}$ and the predicted trajectory $\underline{x}_\textrm{pred}$ is smaller than a threshold $\underline{\Delta}$ in each coordinate.
For each coordinate, we calculate the maximal time the error is below the threshold with
\begin{equation}
    \underline{v} = \argmax_t \left\{ \left| \, x_i(t) - x_{i,\,\textrm{pred}}(t)\right| < \Delta_i \right\} \;\;\;.
\end{equation}
The $\argmax$ function is applied element-wise.
We use the standard deviation $\sigma$, applied element-wise as the threshold $\underline{\Delta}$ with $\underline{\Delta} = \sigma{\left( \underline{x}\right)}$.

The forecast horizon $v$ is then defined as the minimal time across all coordinates during which the prediction error stays below the threshold by
\begin{equation}
    v = \min \underline{v}\;\;\;.
\end{equation}

In order to compare the forecast horizon across different systems, we define it in multiples of Lyapunov times as
\begin{equation}
    v_\lambda = \frac{v}{\tau_\lambda} = v\,\lambda\;\;\;.
\end{equation}

\section{Illustration of our SINDy library}
In the SINDy framework, the library of all possible nonlinearities is of key importance as it constrains the dynamics of the system to certain terms.
Here, we want to illustrate our definition of the library with a concrete example.
We present our definition using the same configuration as in our experiments. Namely, we use a global list of all possible nonlinearities constructed as fractions with a denominator of $50$ and the numerator ranging from $10$ to $300$ in steps of $10$ as
\begin{equation}
\mathbf{H} = \left\{
\dfrac{10}{50},\,\dfrac{20}{50},\,\dots,\,\dfrac{300}{50}\right\}\;\;\;.
\end{equation}
For a nonlinearity $\eta$, we then use all combinations of terms that can be built from $\mathbf{H}$ and result in that nonlinearity.
\textit{E.g.}, for $\eta=\nicefrac{80}{50}$ the library will contain the nonlinear terms given by
\begin{widetext}
\begin{equation}
\label{equ:sindyexample}
    \mathbf{X}^{\nicefrac{80}{50}} = \begin{pmatrix}
        a^{\nicefrac{80}{50}} & a^{\nicefrac{10}{50}}\,b^{\nicefrac{70}{50}} & a^{\nicefrac{20}{50}}\,b^{\nicefrac{60}{50}} & \cdots & a^{\nicefrac{40}{50}}\,b^{\nicefrac{40}{50}} & a^{\nicefrac{10}{50}}\,b^{\nicefrac{10}{50}}\,c^{\nicefrac{60}{50}} & \cdots & a^{\nicefrac{20}{50}}\,b^{\nicefrac{30}{50}}\,c^{\nicefrac{30}{50}}\\
        \vdots & &  &  & & & & \vdots
    \end{pmatrix}\;\;\;.
\end{equation}
\end{widetext}
Here, we use the variable names $a$, $b$, and $c$ instead of $x_0$, $x_1$, and $x_2$ to visualize that all possible combinations are considered. $a^{\alpha_1}\,b^{\alpha_2}$ represents $x_0^{\alpha_1}\,x_1^{\alpha_2}$, $x_1^{\alpha_1}\,x_0^{\alpha_2}$, $x_0^{\alpha_1}\,x_2^{\alpha_2}$,
$x_2^{\alpha_1}\,x_0^{\alpha_2}$,
$x_1^{\alpha_1}\,x_2^{\alpha_2}$, and finally $x_2^{\alpha_1}\,x_1^{\alpha_2}$. This is repeated analogously for single terms $a^{\alpha_1}$ and terms containing three factors $a^{\alpha_1}\,b^{\alpha_2}\,c^{\alpha_3}$.
We omit the time dependence of the variables in our notation and imply temporal evolution through the progression of the rows as done in the original work\cite{SINDy}.

Using this method the number of possible combinations of nonlinear terms can quickly become very large depending on the spacing within the global nonlinearities $\mathbf{H}$.
In our work, we find a spacing of $\nicefrac{10}{50}$ to be an appropriate trade-off between granularity of the nonlinearities and size of the library.
However, this method is only meant to be a proof of concept. Further in-depth studies may be required to find the optimal setup.

\section*{References}
\bibliography{bibliography}

@techreport{Jaeger2001,
    author = {Herbert Jaeger},
    title = {The ``echo state'' approach to analysing and training recurrent neural networks - with an {E}rratum note},
    institution = {GMD Forschungszentrum Informationstechnik},
    year = {2001}
}

@article{Maass2002,
  title={Real-time computing without stable states: a new framework for neural computation based on perturbations},
  author={Wolfgang Maass and Thomas Natschläger and Henry Markram},
  journal={Neural Comput.},
  volume={14},
  number={11},
  pages={2531-2560},
  year={2002},
}

@article{Jaeger2004,
  title = {{H}arnessing {N}onlinearity: {P}redicting {C}haotic {S}ystems and {S}aving {E}nergy in {W}ireless {C}ommunication},
  author={Herbert Jaeger and Harald Haas},
  journal={Science},
  volume={304},
  number={5667},
  pages={78-80},
  year={2004},
}

@article{Gauthier2021,
  title={Next generation reservoir computing},
  author={Daniel J. Gauthier and Erik Bollt and Aaron Griffith and Wendson A. S. Barbosa},
  journal={Nat. Comm.},
  volume={12},
  pages={5564},
  year={2021}
}

@article{Barbosa22,
    author = {Wendson A. S. Barbosa and Daniel J. Gauthier},
    title = {Learning spatiotemporal chaos using next-generation reservoir computing},
    journal = {Chaos},
    volume = {32},
    number = {9},
    pages = {093137},
    year = {2022}
}

@article{Lorenz1963,
    title = {Deterministic {N}onperiodic {F}low},
    journal = {JAS},
    volume = {20},
    number = {2},
    pages = {130-141},
    year = {1963},
    author = {Edward N. Lorenz}}

@article{Pathak2017,
    title = {Using machine learning to replicate chaotic attractors and calculate {L}yapunov exponents from data},
    journal = {Chaos},
    volume = {27},
    pages = {121102},
    year = {2017},
    author = {Jaideep Pathak and Zhixin Lu and Brian R. Hunt and Michelle Girvan and Edward Ott}}

@article{Lu2018,
    title = {Attractor reconstruction by machine learning},
    journal = {Chaos},
    volume = {28},
    pages = {061104},
    year = {2018},
    author = {Zhixin Lu and Brian R. Hunt and Edward Ott}}

@article{RungeKutta45,
    author = {John R. Dormand and Pete J. Prince},
    title = {A family of embedded {R}unge--{K}utta formulae},
    journal = {J. Comput. Appl. Math.},
    year = {1980},
    pages = {19-26},
    volume = {6},
    number = {1}
}

@article{SCAN,
    author = {Sebastian Baur and Tamon Nakano and Dennis Duncan and Fabian Fischbach and Alexander Haluszczynski and Micheal Klatt and Daniel Köglmayr and Haochun Ma and Davide Prosperino and Christoph Räth},
    title = {{SCAN}: A versatile implementation of reservoir computing methods},
    journal = {\textit{(in preparation)}},
    doi = {https://github.com/DLR-KI/scan}
}

@book{Sprott2010Halvorsen,
    place={Singapore},
    title={Elegant Chaos},
    publisher={World Scientific Publishing},
    author={Julien C. Sprott},
    year={2010}}

@article{Ma2023MinimalRC,
    author = {Haochun Ma and Davide Prosperino and Christoph Räth},
    title = {A novel approach to minimal reservoir computing},
    journal = {Sci. Rep.},
    year = {2023},
    volume = {13},
    pages = {12970}
}

@article{Ma2022Causality,
    author = {Haochun Ma and Alexander Haluszczynski and Davide Prosperino and Christoph Räth},
    title = {Identifying causality drivers and deriving governing equations of nonlinear complex systems},
    journal = {Chaos},
    year = {2022},
    volume = {32},
    pages = {103128}
}

@article{Raeth2012,
  title = {Revisiting {A}lgorithms for {G}enerating {S}urrogate {T}ime {S}eries},
  author = {Christoph Räth and Mario Gliozzi and Iossif E. Papadakis and Wolfgang Brinkmann},
  journal = {Phys. Rev. Lett.},
  volume = {109},
  issue = {14},
  pages = {144101}
}

@article{Theiler1992,
title = {Testing for nonlinearity in time series: the method of surrogate data},
journal = {Physica D},
volume = {58},
number = {1},
pages = {77-94},
author = {James Theiler and Stephen Eubank and André Longtin and Bryan Galdrikian and James {Doyne Farmer}}
}

@article{LukoseviciusJaeger2009,
    author = {Mantas Lukoševičius and Herbert Jaeger},
    title = {Reservoir computing approaches to recurrent neural network training},
    journal = {Comput. Sci. Rev.},
    year = {2009},
    pages = {127-149},
    volume = {3}
}

@article{Haluszczynski2019,
    author = {Alexander Haluszczynski and Christoph Räth},
    title = {Good and bad predictions: {A}ssessing and improving the replication of chaotic attractors by means of reservoir computing},
    journal = {Chaos},
    year = {2019},
    pages = {103143},
    volume = {29}
}

@article{RidgeRegression,
    author = {Arthur E. Hoerl and Robert W. Kennard},
    title = {{R}idge {R}egression: {A}pplications to {N}onorthogonal {P}roblems},
    journal = {Technometrics},
    year = {1970},
    volume = {12},
    number = {1},
    pages = {69-82}
}

@article{Ma2023Blockdiagnoal,
    author = {Haochun Ma and Davide Prosperino and Alexander Haluszczynski and Christoph Räth},
    title = {Efficient forecasting of chaotic systems with block-diagonal and binary reservoir computing},
    journal = {Chaos},
    year = {2023},
    pages = {063130},
    volume = {33}
}

@article{Li2024,
    author = {Xin Li and Qunxi Zhu and Chengli Zhao and Xiajun Duan and Bolin Zhao and Xue Zhang and Huanfei Ma and Jie Sun and Wei Lin},
    title = {Higher-order {G}ranger reservoir computing: simultaneously achieving scalable complex structures inference and accurate dynamics prediction},
    journal = {Nat. Comm.},
    year = {2024},
    pages = {2506},
    volume = {15}
}

@article{Thomas1999,
    author = {René Thomas},
    title = {Deterministic chaos seen in terms of feeedback circuits: analysis, synthesis, ``labyrinth chaos''},
    journal = {Int. J. Bifurc. Chaos},
    year = {1999},
    pages = {1889-1905},
    volume = {9}
}

@article{Herteux2020,
    author = {Joschka Herteux and Christoph Räth},
    title = {Breaking symmetries of the reservoir equations in echo state networks},
    journal = {Chaos},
    year = {2020},
    volume = {30},
    pages = {123142}
}

@article{Stepney2024,
    author = {Susan Stepney},
    title = {Physical reservoir computing: a tutorial},
    journal = {Nat. Comput.},
    year = {2024},
    volume = {23},
    pages = {665-685}
}

@article{Fama1965,
    author = {Eugene F. Fama},
    title = {The {B}ehavior of {S}tock-{M}arket {P}rices},
    journal = {J. Bus.},
    year = {1965},
    volume = {38},
    pages = {34-105}
}

@article{Rosenstein1993,
    author = {Micheal T. Rosenstein and James J. Collins and Carlo J. {De Luca}},
    title = {A practical method for calculating largest {L}yapunov exponents from small data sets},
    journal = {Physica D},
    year = {1993},
    volume = {65},
    pages = {117-134}
}

@article{Grassberger1983,
    author = {Peter Grassberger and Itamar Procaccia},
    title = {Measuring the {S}trangeness of {S}trange {A}ttractors},
    journal = {Physica D},
    year = {1983},
    volume = {9},
    pages = {189-208}
}

@incollection{Maneewongvatana2002,
    author = {Songrit Maneewongvatana and David M. Mount} ,
    title = {Analysis of {A}pproximate {N}earest {N}eighbor {S}earching with {C}lustered {P}oint {S}ets},
    series = {{DIMACS} {S}eries in {D}iscrete {M}athematics and {T}heoretical {C}omputer {S}cience},
    booktitle = {Data {S}tructures, {N}ear {N}eighbor {S}earches, and {M}ethodology: {F}ifth and {S}ixth {DIMACS} {I}mplementation {C}hallenges},
    volume = {59},
    year = {2002},
    pages = {105-123},
    editor = {Micheal H. Goldwasser and David S. Johnson and Catherine C. McGeoch},
    publisher = {American Mathematical Society}
}

@article{Shahi2021,
	title        = {Long-{T}ime {P}rediction of {A}rrhythmic {C}ardiac {A}ction {P}otentials {U}sing {R}ecurrent {N}eural {N}etworks and {R}eservoir {C}omputing},
	author       = {Shahrokh Shahi and Christopher D. Marcotte and Conner J. Herndon and Flavio H. Fenton and Yohannes Shiferaw and  Elizabeth M. Cherry},
	year         = 2021,
	journal      = {Front. Physiol.},
	volume       = 12,
	pages        = 734178
}

@article{Brucke2024,
	title        = {Benchmarking reservoir computing for residential energy demand forecasting},
	author       = {Karoline Brucke and Simon Schmitz and Daniel Köglmayr and Sebastian Baur and Christoph Räth and and Esmail Ansari and Peter Klement},
	year         = {2024},
	journal      = {Energy Build.},
	volume       = {314},
	pages        = {114236}
}

@article{Herteux2024,
	title        = {Forecasting trends in food security with real time data},
	author       = {Joschka Herteux and Christoph Räth and Giulia Martini and Amine Baha and Kyriacos Koupparis and Ilaria Lauzana and Duccio Piovani},
	year         = {2024},
	journal      = {Commun. Earth Environ.},
	volume       = {5},
	pages        = {611}
}

@article{Carroll2019,
	title        = {Network structure effects in reservoir computers},
	author       = {Thomas L. Carroll and Louis M. Pecora},
	year         = {2019},
	journal      = {Chaos},
	volume       = {29},
	pages        = {083130}
}

@article{Mijalkov2025,
	title        = {Computational memory capacity predicts aging and cognitive decline},
	author       = {Mite Mijalkov and Ludvig Storm and Blanca Zufiria-Gerbolés and Dániel Veréb and Zhilei Xu and Anna Canal-Garcia and Jiawei Sun and Yu-Wei Chang and Hang Zhao and Emiliano Gómez-Ruiz and Massimiliano Passaretti and Sara Garcia-Ptacek and Miia Kivipelto and Per Svenningsson and Henrik Zetterberg and Heidi Jacobs and Kathy Lüdge and Daniel Brunner and Bernhard Mehlig and Giovanni Volpe and Joana B. Pereira},
	year         = {2025},
	journal      = {Nat. Comm.},
	volume       = {16},
	pages        = {2748}
}

@article{Ohkubo2024,
	title        = {Reservoir computing with generalized readout based on generalized synchronization},
	author       = {Akane Ohkubo and Masanobu Inubushi},
	year         = {2024},
	journal      = {Sci. Rep.},
	volume       = {14},
	pages        = {30918}
}

@article{AMOC2023,
	title        = {Warning of a forthcoming collaps of the {A}tlantic meridional overturning circulation},
	author       = {Peter Ditlevsen and Susanne Ditlevsen},
	year         = {2023},
	journal      = {Nat. Comm.},
	volume       = {14},
	pages        = {4254}
}

@article{SINDY,
	title        = {Discovering governing equations from data by sparse identification of nonlinear dynamical systems},
	author       = {Steven L. Brunton and Joshua L. Proctor and J. Nathan Kutz},
	year         = {2016},
	journal      = {PNAS},
	volume       = {113},
	pages        = {3932-3937}
}

@article{Laut2016,
	title        = {Surrogate-assisted network analysis of nonlinear time series},
	author       = {Ingo Laut and Christoph Räth},
	year         = {2016},
	journal      = {Chaos},
	volume       = {26},
	pages        = {103108}
}

@article{Schreiber2018,
	title        = {Phase walk analysis of leptokurtic time series},
	author       = {Korbinian Schreiber and Heike I. Modest and Christoph Räth},
	year         = {2018},
	journal      = {Chaos},
	volume       = {28},
	pages        = {063120}
}

@article{Bhadriraju2020,
author = {Bhavana Bhadriraju and Mohammed Saad Faizan Bangi and Abhinav Narasingam and Joseph Sang-Il Kwon},
title = {Operable adaptive sparse identification of systems: {A}pplication to chemical processes},
journal = {AIChE J.},
volume = {66},
pages = {e16980},
year = {2020}
}

@article{Paparazzo2025,
author = {Francesco Paparazzo and Andrea Castoldi and Mohammed Irshadh Ismaaeel Sathyamangalam Imran and Stefano Arrigoni and Francesco Braghin},
title = {Learning-{B}ased {MPC} {L}everaging {SINDy} for {V}ehicle {D}ynamics {E}stimation},
journal = {Electronics},
volume = {14},
pages = {1935},
year = {2025}}

@article{pysindy1,
    author = {Brian M. {de Silva} and Kathleen Champion and Markus Quade and Jean-Christophe Loiseau and J. Nathan Kutz and Steven L. Brunton},
    title = {{PySINDy}: {A} {P}ython package for the sparse identification of nonlinear dynamical systems from data},
    journal = {J. Open Source Softw.},
    volume = {5},
    year = {2020},
    pages = {2104}
}

@article{pysindy2,
year = {2022},
volume = {7},
pages = {3994},
author = {Alan A. Kaptanoglu and Brian M. de Silva and Urban Fasel and Kadierdan Kaheman and Andy J. Goldschmidt and Jared Callaham and Charles B. Delahunt and Zachary G. Nicolaou and Kathleen Champion and Jean-Christophe Loiseau and J. Nathan Kutz and Steven L. Brunton},
title = {{PySINDy}: A comprehensive {P}ython package for robust sparse system identification},
journal = {J. Open Source Softw.},
}

@article{Prosperino2025,
year = {2025},
volume = {6},
pages = {015012},
author = {Davide Prosperino and Haochun Ma and Christoph Räth},
title = {A generalized method for estimating parameters of chaotic systems using synchronization with modern optimizers},
journal = {J. Phys. Complex.},
}

@article{erdosrenyi,
year = {1960},
volume = {5},
pages = {17-61},
author = {Paul Erdős and Alfréd Rényi},
title = {On the {E}volution of {R}andom {G}raphs},
journal = {Publ. Math. Inst. Hung. Acad. Sci},
}

@article{Cont2001,
year = {2001},
volume = {1},
pages = {223-236},
author = {Rama Cont},
title = {Empirical properties of asset returns: stylized facts and statistical issues},
journal = {Quant. Finance},
}

@article{Zhang2023,
year = {2023},
volume = {5},
pages = {033213},
author = {Yuanzhao Zhang and Sean P. Cornelius},
title = {Catch-22s of reservoir computing},
journal = {Phys. Rev. Res.},
}

\end{document}